\documentclass[11pt]{article}

\usepackage[final]{acl}

\usepackage{times}
\usepackage{latexsym}

\usepackage[T1]{fontenc}

\usepackage[utf8]{inputenc}

\usepackage{microtype}

\usepackage{inconsolata}

\usepackage{graphicx}

\usepackage{amsmath}
\usepackage{amssymb}
\usepackage{mathtools}
\usepackage{amsthm}
\usepackage{amsfonts}
\usepackage{booktabs}
\usepackage{multirow}
\usepackage{caption}
\usepackage{subcaption}
\usepackage{longtable}
\usepackage{tabularx}
\usepackage{arydshln}
\usepackage{enumitem}
\usepackage{hyperref}
\usepackage{soul}
\usepackage{siunitx}    

\usepackage{pifont}
\usepackage{dsfont}
\usepackage[most]{tcolorbox}

\usepackage[textsize=tiny]{todonotes}

\definecolor{lightyellow}{rgb}{1, 1, 0.85}
\definecolor{lightblue}{rgb}{0.85, 1, 1}

\usepackage[table,xcdraw]{xcolor}

\usepackage{CJKutf8}

\newcommand{\xmark}{\ding{55}}%

\newtcolorbox{promptbox}[1]{
    breakable,
    before skip=1.2em,
    after skip=1.2em,
    width=\linewidth,
    colback=white,
    colframe=black,
    boxrule=0.5pt,
    sharp corners,
    fonttitle=\bfseries\sffamily\small,
    colbacktitle=gray!20,
    coltitle=black,
    title=#1,
    fontupper=\sffamily\small,
    left=8pt,
    right=8pt,
    top=8pt,
    bottom=8pt
}

\title{DialToM: A Theory of Mind Benchmark for Forecasting State-Driven Dialogue Trajectories}

\author{
    \textbf{Neemesh Yadav\textsuperscript{1}\thanks{Corresponding Authors}},
    \textbf{Palakorn Achananuparp\textsuperscript{1}\footnotemark[1]},
    \textbf{Jing Jiang\textsuperscript{2}},
    \textbf{Ee-Peng Lim\textsuperscript{1}}
\\
    \textsuperscript{1}Singapore Management University \quad
    \textsuperscript{2}Australian National University
\\
    \texttt{\{neemeshy, palakorna\}@smu.edu.sg},  
    \texttt{jing.jiang@anu.edu.au}, 
    \texttt{eplim@smu.edu.sg}
}

\begin{document}
\maketitle
\begin{abstract}
We introduce DialToM, an annotated Theory of Mind (ToM) benchmark built from naturalistic human--human dialogues using a multiple-choice evaluation framework. Concurrent with recent work showing a gap between explicit mental-state inference and applied ToM in synthetic settings~\cite{gu2024simpletom}, we establish a stricter \emph{State-Driven Diagnostic Probe} in which models must forecast state-consistent dialogue trajectories solely from isolated mental-state profiles without dialogue context. Our evaluation reveals a systematic reasoning asymmetry---LLMs excel at inferring mental states (Literal ToM) but struggle to leverage them for social forecasting (Functional ToM). Crucially, a domain expert achieves 100\% accuracy on this task, proving its validity and establishing a stark human-AI capability gap. Further, a teacher-student reasoning injection probe shows that Gemini 3 Pro---which establishes the leading baseline---possesses robust Functional ToM capabilities for context-free forecasting that are transferable to weaker models. DialToM, its evaluation code, and dataset are publicly available at \url{https://github.com/Stealth-py/DialToM}.
\end{abstract}

\section{Introduction}
Theory of Mind (ToM) is a foundation of human cognition that allows us to reason about others' hidden mental states, such as their Beliefs, Desires, and Intentions (the BDI model) \cite{Bratman1987-ic, Apperly2009}. Human social intelligence relies on a dual-stage execution of this capability: (i) inferring latent mental states from context; and (ii) utilizing those mental states to plan downstream actions \cite{Ho2022, riemer2025position}. Driven by the rapid progress of LLMs, existing research has examined machine ToM in various scenarios. Existing evaluations, however, remain heavily skewed toward the first stage---tracking states and identifying false beliefs in synthetic narrative domains \cite{strachan2024testing,le-etal-2019-revisiting, wang2025rethinkingtheorymindbenchmarks}---leaving open whether LLMs can genuinely move from passive inference to functional planning in naturalistic interactions.

A growing body of work suggests this \textit{inference-to-action} gap is a fundamental limitation of current LLMs. Recently, \citet{gu2024simpletom} and \citet{li2025rectombenchmarkevaluatingmachine} demonstrated that while frontier models reliably infer mental states, this competence fails to translate into downstream behavior prediction. We converge on this asymmetry with \textbf{DialToM}, a human-verified benchmark over naturalistic conversations that operationalizes the distinction between \textbf{Literal ToM} and \textbf{Functional ToM}\footnote{See Appendix~\ref{appendix:literal_functional_tom} for a detailed theoretical background on the distinction between Literal and Functional ToM.} \cite{riemer2025position}. DialToM extends prior investigations along three critical axes: (i) grounding evaluation in high-stakes human dialogues (counseling and persuasion); (ii) adopting a richer six-attribute model featuring a novel \emph{Trust} dimension to capture relational rapport; and (iii) enforcing a bifurcated evaluation that pairs a context-driven \textbf{Retrospective task} (Literal ToM: inferring states from dialogue) with a strictly context-free \textbf{Prospective Diagnostic Probe} (Functional ToM: forecasting actions from isolated mental states).

\begin{figure*}[t]
    \centering
    \includegraphics[width=0.85\linewidth]{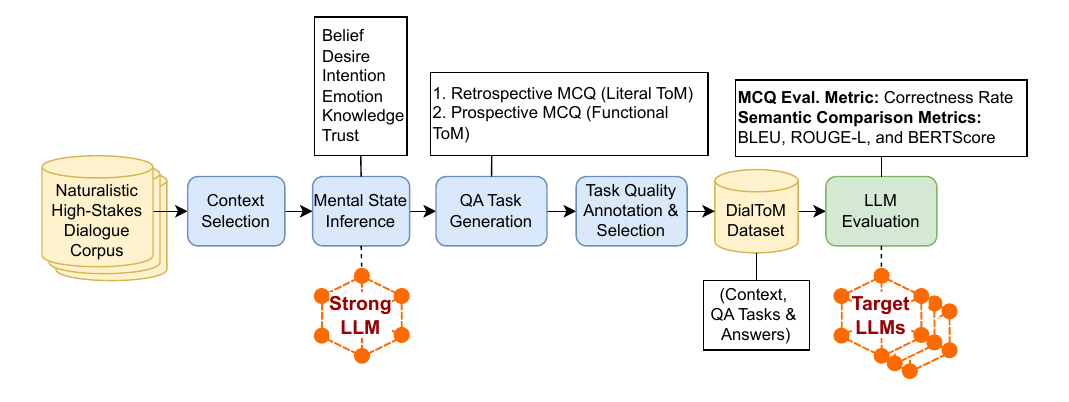}
    \caption{The DialToM Multiple-Choice Question (MCQ) Evaluation Benchmarking Pipeline. The workflow illustrates the transition from Retrospective Inference to Prospective Diagnostic Forecasting through automated mental state inference, distractor generation, and subsequent human verification.}
    \label{fig:framework}
\end{figure*}

The DialToM evaluation pipeline (Figure~\ref{fig:framework}) benchmarks models across both the Retrospective inference task and a context-free Prospective diagnostic forecasting task. Within this framework, models must identify state-consistent dialogue trajectories based solely on an isolated mental-state profile---a setup capturing the literal, high-fidelity internal monologue and communicative intent that directly dictate a speaker's verbal choices. This context-free setup is grounded in standard psychometric principles, which dictate that validating a cognitive construct requires evaluating the bidirectional mapping between internal states and observed behaviors \cite{cronbach1955construct, whitely1983construct}. The strict design is validated by a professional domain expert who achieved 100\% accuracy under these exact conditions, and by the task structure itself\footnote{See Appendix~\ref{appendix:task_examples} for concrete examples.}, where candidate trajectories share high surface-form similarity, forcing a reliance on genuine mental-state reasoning.

Our empirical analysis reveals a systematic \emph{reasoning asymmetry}. While models excel at Retrospective mental state inference, the majority fail at Prospective Diagnostic Forecasting, with several scoring near or below chance. This asymmetry is consistent across domains and model families, replicating and deepening the inference-to-action gap identified by \citet{gu2024simpletom, li2025rectombenchmarkevaluatingmachine}. In addition, we find that LLMs specifically struggle with the Trust attribute, suggesting relational dynamics remain an understudied dimension. A notable exception is Gemini 3 Pro, which achieves $\approx$83\% on the Prospective task---and whose reasoning traces, when applied to weaker models, boost their performance by up to 76 points. This establishes an empirical proof of concept for functional ToM-grounding in LLMs.

\section{Related Work}

\begin{table*}[t]
    \centering
    \resizebox{0.95\linewidth}{!}{
    \begin{tabular}{l c c c c c}
    \toprule
        \textbf{Benchmark} & \textbf{Attributes} & \textbf{Context Type} & \textbf{Functional ToM} & \textbf{Context-Free Prediction} & \textbf{Context Source} \\ \midrule
        ToMi \cite{le-etal-2019-revisiting} & B & Goal-oriented stories & \xmark & \xmark & Template-based \\
        MindCraft \cite{bara2021mindcraft} & BIK & Situated collaborative dialogue & \xmark & \xmark & Human-Human \\
        CPA \cite{bara2023towards} & BIK & Situated collaborative dialogue & \xmark & \xmark & Human-Human \\
        FANToM \cite{kim-etal-2023-fantom} & B & Synthetic conversations & \xmark & \xmark & LLM generated \\
        Hi-ToM \cite{wu-etal-2023-hi} & B & Goal-oriented stories & \xmark & \xmark & Template-based \\
        Common-ToM \cite{soubki2024views} & B & Natural spoken dialogue & \xmark & \xmark & Human-Human \\
        OpenToM \cite{xu-etal-2024-opentom} & B & Goal-oriented stories & \xmark & \xmark & LLM generated \\
        ToMBench \cite{chen-etal-2024-tombench} & BDIEK & Goal-oriented stories & \xmark & \xmark & Human-written \\
        DynToM \cite{xiao-etal-2025-towards} & BIE & Synthetic scenarios & \xmark & \xmark & LLM generated \\
        ToMATO \cite{shinoda2025tomato} & BDIEK & Social simulation & \xmark & \xmark & LLM-LLM \\
        \midrule
        BigToM \cite{kanishk2023bigtom} & B & Goal-oriented stories & \checkmark & \xmark & Template-based \\
        SimpleToM \cite{gu2024simpletom} & B & Synthetic stories & \checkmark & \xmark & LLM+Human \\
        RecToM \cite{li2025rectombenchmarkevaluatingmachine} & BDI & Recommendation dialogue & \checkmark & \xmark & Human-Human \\
        \midrule \midrule
        \textbf{DialToM (Ours)} & \textbf{BDIEK-T} & \textbf{High-stakes multi-turn dialogues} & \checkmark & \checkmark & \textbf{Human-Human} \\
    \bottomrule
    \end{tabular}}
    \caption{Comparison of DialToM with existing Theory of Mind benchmarks. \textbf{Functional ToM} indicates whether the benchmark measures applied action/trajectory forecasting based on mental state profiles. \textbf{Context-Free Prediction} indicates whether this forecasting task strips out conversational or story context. DialToM uniquely operates at the intersection of both properties while analyzing high-stakes dialogues with Trust modeling.}
    \label{tab:comparison_benchmarks}
\end{table*}

\subsection{Benchmarking Theory of Mind in LLMs}
\paragraph{Literal ToM \& Its Limitations}
Early ToM benchmarks adapted classic false-belief paradigms \cite{Dennett_1978, Wimmer1983} into rigid, template-based narratives (ToMi \cite{le-etal-2019-revisiting}, Hi-ToM \cite{wu-etal-2023-hi}, OpenToM \cite{xu-etal-2024-opentom}), later extended to multi-agent simulations like FANToM \cite{kim-etal-2023-fantom}, SOTOPIA \cite{zhou2024sotopia}, ToMBench \cite{chen-etal-2024-tombench}, and ToMATO \cite{shinoda2025tomato}, and to naturalistic, situated dialogue via Common-ToM \cite{soubki2024views}, MindCraft \cite{bara2021mindcraft}, and its collaborative plan acquisition extension \cite{bara2023towards}. Yet these remain confined to passive, retrospective state attribution and fragile to superficial perturbations \cite{ullman2023large}, overlooking the broader spectrum of real-world social cognition \cite{wang2025rethinkingtheorymindbenchmarks, xiao-etal-2025-towards}.
\paragraph{Functional ToM \& The Inference-to-Action Gap}
To bridge this gap, recent benchmarks pioneer behavioral prediction tasks for Functional ToM. BigToM \cite{kanishk2023bigtom} forecasts behavior in synthetic narratives, while RecToM \cite{li2025rectombenchmarkevaluatingmachine} and SimpleToM \cite{gu2024simpletom} jointly probe state attribution and behavior prediction, both documenting an inference-to-action gap---\emph{cognitive inference vs. behavioral prediction} and \emph{Explicit vs. Applied ToM}, respectively---where frontier LLMs track mental states but fail to act on them. All rely on context-driven setups retaining full conversational history. DialToM instead combines context-free prediction with naturalistic human--human dialogue and a Trust dimension, occupying a position no existing resource addresses (Table~\ref{tab:comparison_benchmarks}).

\subsection{Distinction of DialToM}
DialToM distinguishes itself from current benchmarking methodologies by evaluating ToM over natural human--human conversations and incorporating a newly proposed Trust dimension between speakers. More importantly, we operationalize \emph{Context-Free Prediction}---the ability to identify state-consistent dialogue trajectories based strictly on an isolated mental state profile. By deliberately withholding prior conversation history, we prevent models from relying on spurious context-to-dialogue correlations.

In Table~\ref{tab:comparison_benchmarks}, we compare DialToM with existing benchmarks across five dimensions: the mental state attributes of focus, the type of contexts used, whether the benchmark measures Functional ToM, whether it utilizes Context-Free Prediction, and the source of the contexts.

\paragraph{Ecological Validity \& The Trust Dimension}
Most benchmarks either focus on goal-oriented synthetic stories or do not follow naturalistic contexts. Recent naturalistic benchmarks like Common-ToM, MindCraft, and CPA mark progress here, but draw from casual or task-oriented interaction rather than the high-stakes, relationally charged settings where trust is most consequential. DialToM fills this gap by drawing from naturalistic, high-stakes human dialogues (mental health support and persuasion). Existing benchmarks, naturalistic or otherwise, do not capture the relational trust dynamics central to these domains. We extend the standard mental state model to include a new Trust (T) component, crucial in human interaction.

\paragraph{Context-Free Prediction} 
We introduce context-free prediction (state-driven probing) to rigorously audit the functional utility of ToM. This diagnostic probe evaluates a model's ability to forecast actions without access to scenario or dialogue history. Unlike context-driven forecasting in prior benchmarks, DialToM challenges models to perform strict diagnostic mapping from isolated mental profiles to logical dialogue trajectories.  This isolation forces models to demonstrate functionally grounded ToM reasoning rather than exploiting surface-level topic coherence.

\paragraph{Supplementary Semantic Verification}
While DialToM's core diagnostic relies on multiple-choice evaluation, it is also uniquely supplemented by a gold-standard reference set of human-written mental state rationales. This allows for the semantic comparison of LLM-generated and human-written reasoning traces, providing an additional qualitative lens for evaluating machine ToM alignment.

\section{Methodology}

\subsection{Data Curation \& Token Normalization}
To ensure ecological validity, DialToM is built upon naturalistic, high-stakes human interactions drawn from three relational domains\footnote{Appendix~\ref{sec:appendix_data_sources} provides complete corpus descriptions and full pre-/post-segmentation statistics (Table~\ref{tab:full_general_stats}).}: Motivational Interviewing via \textbf{AnnoMI}~\cite{annomi2022}, Emotional Support Conversations via \textbf{ESConv}~\cite{liu-etal-2021-towards}, and strategic negotiation via \textbf{PersuasionForGood} (PFG)~\cite{wang-etal-2019-persuasion}. We preserve a constant dyadic role distribution, mapping participants to an \emph{initiator} (counselor, supporter, persuader) and a \emph{recipient} (client, seeker, persuadee).

To normalize context density and mitigate extreme length variance across domains (ranging from an average of 1,233 words in AnnoMI to 273 words in PFG), we segment full sessions into uniform sliding context windows. We enforce a strict reasoning horizon bounded between 150 and 200 words per window to optimize semantic completeness and token efficiency. Any trailing session fragments falling below the 150-word threshold are dynamically concatenated with their preceding window to maintain interaction continuity.

As summarized in Table~\ref{tab:general_stats}, this pipeline yields 5,943 evaluation contexts. We generate 6 Retrospective questions (one per BDIEKT attribute) and 1 Prospective question per context, establishing a comprehensive benchmarking suite of 35,658 Literal ToM tasks and 5,943 Functional ToM forecasting problems.

\begin{table}[t] 
    \centering
    \resizebox{\linewidth}{!}{
    \begin{tabular}{l rrrr}
    \toprule
         \textbf{Statistic} & \textbf{MI} & \textbf{ESC} & \textbf{PFG} & \textbf{Overall} \\ \midrule
         Context Count & 865 & 3,215 & 1,863 & 5,943 \\
         Word Count per Context & 188.2 & 205.0 & 210.8 & — \\
         Turn Count per Context & 5.3 & 4.5 & 5.4 & — \\
         Word Count per Turn & 35.5 & 45.5 & 39.1 & — \\ \midrule
         Retrospective Infer. Qns. & 5,190 & 19,290 & 11,178 & 35,658 \\
         Prospective Forecast. Qns. & 865 & 3,215 & 1,863 & 5,943 \\
         Human Expert Gold Refs. & 300 & 300 & 300  & 900\\
     \bottomrule
    \end{tabular}}
    \caption{General descriptive statistics of the curated DialToM evaluation suite across its three core dialogue domains.}
    \label{tab:general_stats}
\end{table}

\subsection{The Trust Dimension in Mental State Modeling}
Previous Theory of Mind (ToM) benchmarks overwhelmingly evaluate five foundational cognitive components: Beliefs (B), Desires (D), Intentions (I), Emotions (E), and Knowledge (K). However, these attributes fail to capture the relational complexity and evolving rapport of naturalistic human interactions. To address this gap, we introduce a novel sixth mental state attribute: \textbf{Trust (T)}.

Trust builds incrementally in multi-turn dialogue and serves as the foundation for collaboration and persuasion. While organizational psychology operationalizes trust via three distinct dimensions—ability, benevolence, and integrity~\cite{Mayer1995}—the DialToM framework unifies these factors into a single concept reflecting active interpersonal rapport strength. This integration is grounded in literature establishing trust as a critical metric in communication and high-stakes social interactions~\cite{Blacksher2016, clark2019whatmakes, Hancock2023}. Modeling Trust alongside standard BDIEK attributes allows DialToM to comprehensively audit how LLMs reason about the interpersonal relational dynamics that govern human interaction.

\subsection{The Evaluation Tasks}
For each dialogue session, the DialToM benchmark consists of two distinct multiple-choice Question-Answering (QA) tasks featuring a single ground-truth answer and three plausible, model-generated distractors:\footnote{All candidate options are generated by GPT-4o (prompt in Appendix~\ref{appendix:prompts}) and rigorously filtered via our human-verification pipeline.}

\begin{enumerate}[noitemsep, leftmargin=*]
    \item \textbf{Retrospective Inference}: The model is provided with the multi-turn conversation history up to turn $i$ ($\mathcal{H}_i$) as its context and must explicitly infer the specific mental state attribute $s$ of the target speaker. This retrospective inference measures the model's capacity for conventional mental state attribution:
    \begin{gather*}
        \mathcal{S}_i = \{B_i, D_i, I_i, E_i, K_i, T_i\} \\
        \mathcal{O}_{R,i} = \{s_i, d^1_{s,i}, d^2_{s,i}, d^3_{s,i}\} \\
        \hat{s}_i = \arg\max_{s' \in \mathcal{O}_{R,i}} \mathcal{M}(s' \mid \mathcal{H}_i) \;\; \forall s\!\in\!\mathcal{S}_i,\, i\!\le\!|\mathcal{H}|
    \end{gather*}
    where $\mathcal{S}_i$ represents the set of mental states (Belief, Desire, Intention, Emotion, Knowledge, Trust), $\mathcal{O}_{R,i}$ is the option space containing ground-truth $s_i$ and three distractor labels $d_{s,i}$, and $\hat{s}_i$ is the value predicted by model $\mathcal{M}$.

    \item \textbf{Prospective Diagnostic Forecasting}: The model is provided strictly with the complete, isolated mental-state profile $\mathcal{S}_i$ and must identify the most state-consistent downstream dialogue trajectory. Vitally, the dialogue history $\mathcal{H}_i$ is entirely withheld---a deliberate design choice that eliminates lexical shortcut exploitation and forces a reliance on Functional ToM.\footnote{We provide an extended discussion on the methodological trade-offs, task artificiality, and human-expert validation of this context-free configuration in Appendix~\ref{appendix:literal_functional_tom}.} Providing dialogue history allows models to exploit topic coherence rather than engage in true functional-ToM reasoning.\footnote{Our dialogue context degradation analysis (Section~\ref{appendix:context_degradation}) confirms this vulnerability directly, demonstrating that model performance increases monotonically as conversational context is restored.} Formally, we define the diagnostic probe as:
    \begin{gather*}
        \mathcal{O}_{P,i} = \{\mathcal{H}_{i+1}, \mathcal{D}^1_{i+1}, \mathcal{D}^2_{i+1}, \mathcal{D}^3_{i+1}\} \\
        \hat{\mathcal{H}}_{i+1} = \arg\max_{\mathcal{H}' \in \mathcal{O}_{P,i}} \mathcal{M}(\mathcal{H}' \mid \mathcal{S}_i) \quad \forall i \in [1, |\mathcal{H}|]
    \end{gather*}
    where $\mathcal{O}_{P,i}$ represents the candidate option space consisting of the ground-truth trajectory $\mathcal{H}_{i+1}$ alongside three adversarial distractors $\mathcal{D}$, and $\hat{\mathcal{H}}_{i+1}$ is the trajectory selected by $\mathcal{M}$.
\end{enumerate}

\subsection{Human-in-the-Loop Verification}
We employ a human-in-the-loop pipeline to audit all model-generated multiple-choice items and establish robust baselines. A group of 22 professional verifiers\footnote{Platforms, compensation, and worker filtering protocols are detailed in Appendix~\ref{appendix:annotation}.} audited all Retrospective and Prospective task choices to ensure that the ground-truth options were uniquely correct and that the distractors were contextually plausible yet logically incorrect. Separately, a cohort of 9 qualified crowd workers authored open-ended mental state inferences across a subset of 300 contexts to serve as a gold-standard semantic reference.

\paragraph{Human Benchmarking \& Expert Ceiling} To quantify task difficulty, three crowd workers evaluated a subset of 900 Retrospective items (300 per domain) to establish a baseline. This evaluation was restricted to Retrospective tasks due to the high cognitive load of the Prospective setup. To validate construct validity and establish an expert performance ceiling, a professional counselor and an expert co-author completed a sample of 20 context-free Prospective task items spanning all model-difficulty tiers, stratified by model failure rates.\footnote{see Appendix~\ref{subsec:human_benchmarking} for details.}

\subsection{Label Aggregation}
We finalize true task labels by aggregating annotations using the Dawid-Skene (DS) algorithm~\cite{Dawid1979}, fitting models independently across annotation batches. To evaluate inter-annotator agreement, we report Gwet’s AC1~\cite{Gwet2008} due to its stability against the kappa paradox in skewed distributions (Appendix~\ref{appendix:label_agg}). The Retrospective task exhibited exceptionally high consensus ($\text{AC1} \in [0.88, 0.92]$), yielding an 83.25\% dataset retention rate. Conversely, the Prospective task yielded lower agreement ($\text{AC1} \in [0.45, 0.68]$) and a 54.3\% retention rate. More importantly, the lower Prospective retention serves as a vital quality signal. Our adversarial distractor pipeline was intentionally engineered to maintain surface-level semantic similarity to the ground-truth trajectory. This high-difficulty design purposefully filters out borderline contexts where average verifiers disagree. 

Despite this high threshold, the filtered subset remains entirely clear and unambiguous to domain specialists, who achieved a near perfect performance ceiling on these retained items.\footnote{Details are provided in Appendix~\ref{appendix:annotation}.} By prioritizing quality over dataset volume, we ensure that every remaining item acts as a strict Functional-ToM probe. While this aggressive filtering yields a smaller final evaluation set, this scale remains robust for zero-shot LLM benchmarking and is  sufficiently powered to expose systematic reasoning deficits without ambiguous annotations.

\section{Experiments}
\label{sec:experiments}

\subsection{Experimental Setup}
\paragraph{Models} We evaluate ToM capabilities across 14 LLMs\footnote{Details are summarized in Appendix Tables~\ref{tab:model_list} and ~\ref{tab:default_api_parameters}.} spanning diverse parameter scales (12B to 1T parameters), access tiers, and architectures (Standard vs. Mixture-of-Experts). Our evaluation suite includes: \textbf{(a) Closed-source models:} Gemini (3 Pro, 2.5 Pro, 2.5 Flash)~\cite{doshi2025gemini3pro, comanici2025gemini25pushingfrontier}, GPT (4.1, 5)~\cite{gpt41, gpt5}; and \textbf{(b) Open-weights models:} Mistral (Nemo, 3.2 24B)~\cite{mistralnemo2024, mistralsmall32}, Moonshot Kimi K2~\cite{kimiteam2025kimik2openagentic}, Qwen 3 (32B, 235B)~\cite{qwen3technicalreport}, Llama (3.1 8B, 3.3 70B, 4 Maverick)~\cite{grattafiori2024llama3herdmodels, llama4}, DeepSeek-V3~\cite{deepseekai2024deepseekv3technicalreport}, and GPT-OSS 120B~\cite{openai2025gptoss120bgptoss20bmodel}. Open-weights models are queried via OpenRouter, and proprietary models via official provider APIs.

\paragraph{Prompting Strategy} The primary benchmark evaluates models using zero-shot prompting without intermediate token generation.\footnote{Complete prompt templates and examples are provided in Appendix~\ref{appendix:prompts} and~\ref{appendix:task_examples}, respectively.} This ensures the benchmark measures a model's direct capabilities. Internal reasoning traces are permitted for reasoning models, e.g., Gemini 3 Pro, consistent with their intended deployment configuration.

\subsection{Evaluation Metrics}
We evaluate performance across both tasks using the \textbf{Correctness Rate (CR)}, defined as the percentage of correctly answered items:
\begin{equation}
    \text{CR} = \frac{1}{N} \sum_{i=1}^N \mathds{1}[\hat{y}_i = y_i]
\end{equation}
where $N$ is the total number of questions, and $\hat{y}_i, y_i \in \mathcal{O}_i$ represent the model's predicted selection and the ground-truth label from the multiple-choice option space $\mathcal{O}_i = \{A, B, C, D\}$, respectively.

\section{Empirical Analysis}
\label{sec:empirical_analysis}

\subsection{Performance Asymmetry \& Scale Dynamics}
\label{subsec:performance_aymmetry}
As shown in Table~\ref{tab:benchmarking_results}, most LLMs achieve high accuracy ($\ge 80\%$) on the Retrospective task, reliably identifying mental state primitives across domains. Conversely, models consistently fail the Prospective Diagnostic Forecasting task; most score near or below the random baseline ($\le 25\%$). This sharp drop underscores a fundamental reasoning deficit---while models possess the capacity to label underlying mental states, they struggle to predict dialogue trajectories in the absence of local contextual cues. A lone exception is Gemini 3 Pro, which maintains robust Functional ToM capabilities by averaging $\approx 83\%$ accuracy across all three target domains. 

Pearson's Chi-Square tests\footnote{Full statistical reporting is provided in Appendix~\ref{appendix:inference-to-action_gap}.} confirm this Retrospective--Prospective gap is universally significant across all 14 evaluated models and 3 domains ($p < 0.01$). Cram\'er's $V$ effect sizes reveal massive degradation, particularly among frontier models like DeepSeek V3 and GPT-4.1 ($V > 0.80$), proving this is a structural reasoning deficit rather than random variance.

Performance scales consistently with overall model parameters both across and within specific proprietary families. High-capacity frontier models (the Gemini family, Qwen 235B, Kimi K2) serve as the top overall performers. This pattern is mirrored inside isolated families, where Mistral 24B outperforms Mistral Nemo, and GPT-5 outscores GPT-4.1. An isolated anomaly occurs within the Llama family, where the compact Llama 8B surprisingly outperforms its larger counterparts (70B and 4 Maverick).

\begin{table}[t]
\centering
\resizebox{0.9\linewidth}{!}{%
\begin{tabular}{l rrrrrr}
\toprule
\multicolumn{1}{c}{\multirow{2}{*}{\textbf{Model}}} & \multicolumn{3}{c}{\textbf{Retrospective}} & \multicolumn{3}{c}{\textbf{Prospective}} \\ \cmidrule{2-7} 
\multicolumn{1}{c}{} & \multicolumn{1}{c}{MI} & \multicolumn{1}{c}{ESC} & \multicolumn{1}{c}{PFG} & \multicolumn{1}{c}{MI} & \multicolumn{1}{c}{ESC} & \multicolumn{1}{c}{PFG} \\ \midrule
    Human & 32.33 & 24.33 & 24 & - & - & - \\ \hdashline
    Gemini 3 P & 97.71 & 99.15 & 98.78 & 83.09 & 80.46 & 86.96 \\
    GPT-5 & 93.46 & 95.20 & 95.28 & 14.71 & 16.26 & 16.85 \\
    GPT-4.1 & 93.14 & 94.07 & 94.40 & 5.88 & 12.81 & 10.87 \\
    GPT-OSS 120B & 75.49 & 88.98 & 85.25 & 7.35 & 15.76 & 6.52 \\
    Gemini 2.5 P & 92.48 & 97.18 & 94.99 & 38.24 & 29.06 & 19.57 \\
    Gemini 2.5 F & 94.77 & 95.48 & 94.40 & 26.47 & 25.12 & 17.93 \\
    Kimi K2 & 92.16 & 95.48 & 93.22 & 22.06 & 25.62 & 13.04 \\
    DeepSeek v3 & 96.73 & 97.74 & 98.82 & 10.29 & 11.82 & 10.33 \\
    Mistral 24B & 61.11 & 68.64 & 74.34 & 5.88 & 4.93 & 7.6  \\
    Mistral Nemo & 42.16 & 52.26 & 56.64 & 6.62 & 14.29 & 5.44 \\
    Qwen 235B & 95.10 & 97.18 & 97.94 & 17.65 & 29.06 & 17.93 \\
    Llama 70B & 77.45 & 84.75 & 87.91 & 18.42 & 19.70 & 7.07 \\
    Llama 4 & 87.58 & 87.57 & 79.28 & 9.56 & 11.82 & 5.6 \\
    Llama 8B & 30.39 & 27.12 & 34.51 & 6.62 & 16.75 & 10.33 \\
\bottomrule
\end{tabular}%
}
\caption{Model benchmarking results reporting Correctness Rate (CR) across Retrospective and Prospective tasks alongside a non-expert human baseline.}
\label{tab:benchmarking_results}
\end{table}

\subsection{Human Baseline and Expert Ceiling} 
\label{subsec:human_basesline}
On the Retrospective task, the low-performing Lay Participant baseline highlights that non-experts are significantly outperformed by almost all evaluated LLMs\footnote{See Appendix~\ref{subsec:human_error_analysis} for error analysis.}, consistent with prior work~\cite{strachan2024testing}. This reflects the intensive linguistic overhead of parsing specialized, high-stakes dialogue, where human concentration frequently wavers across lengthy annotations. 

On the context-free Prospective task (MI domain; 20 items), a professional counselor and an expert dialogue coder achieved 100\% and 80\% CR scores (95\% CI: [83.2\%, 100\%]), respectively, while Gemini 3 Pro matched the expert baseline with an 80\% CR on those identical items. The near-perfect expert ceiling demonstrates that the context-free formulation remains objectively unambiguous and solvable for domain specialists. The massive performance gap between the expert ceiling and average LLM accuracy ($\le 25\%$) points to a profound functional-ToM reasoning deficit, converging with recent independent findings of an inference-to-action gap in LLMs~\cite{gu2024simpletom, li2025rectombenchmarkevaluatingmachine}.

\subsection{Cross-Domain \& Cross-Task Generalization}
Evaluating cross-domain correlations reveals strong, statistically significant linear connections between all interaction domains across both evaluation paradigms (Retrospective Pearson $r=0.978$; Prospective Pearson $r=0.949$; Table~\ref{tab:cross_domain}). This indicates that a model's primary capability to parse a given ToM dimension generalizes exceptionally well across distinct conversational contexts. In contrast, cross-task correlations are weak and statistically non-significant under a linear assumption ($p = 0.35$), though they display a moderate monotonic trend (Spearman $\rho = 0.67$). This suggests that cross-domain proficiency within a task does not imply cross-task synchronization on DialToM.

\begin{table}[h]
\centering
\resizebox{0.9\linewidth}{!}{%
\begin{tabular}{lllllll} 
\toprule
    \multicolumn{1}{c}{\multirow{2}{*}{\textbf{Domain}}} & \multicolumn{3}{c}{\textbf{Retrospective}} & \multicolumn{3}{c}{\textbf{Prospective}} \\ \cmidrule{2-7}
    \multicolumn{1}{c}{} & \multicolumn{1}{c}{MI} & \multicolumn{1}{c}{ESC} & \multicolumn{1}{c}{PFG} & \multicolumn{1}{c}{MI} & \multicolumn{1}{c}{ESC} & \multicolumn{1}{c}{PFG} \\ \midrule
    MI & 1.0 & 0.978$^\ast$ & 0.965$^\ast$ & 1.0 & 0.966$^\ast$ & 0.941$^\ast$ \\
    ESC & 0.978$^\ast$ & 1.0 & 0.987$^\ast$ & 0.966$^\ast$ & 1.0 & 0.939$^\ast$ \\
    PFG & 0.965$^\ast$ & 0.987$^\ast$ & 1.0 & 0.941$^\ast$ & 0.939$^\ast$ & 1.0 \\
\bottomrule
\end{tabular}%
}
\caption{Pearson correlations across conversation domains. ($^\ast$: p$<$0.001, $^\dagger$: p$<$0.01, $^\ddagger$: p$<$0.05)}
\label{tab:cross_domain}
\end{table}

\subsection{Attribute-Level \& Semantic Analysis}
Fine-grained attribute mapping (Table~\ref{tab:state_dispersed_results}) demonstrates that nearly all evaluated LLMs perform poorly when isolating the newly introduced \emph{Trust} attribute compared to foundational BDI attributes. For instance, high-performing models like Qwen 235B and Gemini 2.5 Pro experience notable performance drops specifically on Trust items. Conversely, superior tracking metrics on Beliefs and Intentions align with prior literature indicating strong modeling biases toward basic factual perspective-taking~\cite{zhu2024language}. These findings validate the inclusion of Trust as an essential but fundamentally underrepresented dimension in current LLM-ToM diagnostics. 

Finally, a complete semantic and structural evaluation comparing LLM outputs directly against open-ended human gold references, incorporating standard NLP metrics (such as BLEU, ROUGE-L, and BERTScore) and an LLM-as-a-Judge protocol (evaluating Propositional Agreement, Causal Attribution, and Specificity Alignment), is deferred to Appendix~\ref{appendix:semantic}. Briefly, while current language models show moderate baseline similarity to human references, they consistently exhibit coarse, low-specificity reasoning compared to human baseline.

\begin{table}[h]
\centering
\resizebox{0.9\linewidth}{!}{%
\begin{tabular}{lrrrrrr}
\toprule
    \multirow{2}{*}{Model} & \multicolumn{6}{c|}{\textbf{Retrospective}} \\ \cmidrule{2-7} 
     & \multicolumn{1}{c}{B} & \multicolumn{1}{c}{D} & \multicolumn{1}{c}{I} & \multicolumn{1}{c}{E} & \multicolumn{1}{c}{K} & \multicolumn{1}{c|}{T} \\ \midrule
    Gemini 3 P & 97.65 & 99.36 & 98.04 & 100 & 98.04 & 98.34 \\
    GPT-5 & 95.40 & 97.47 & 96.10 & 98.86 & 97.42 & 84.15 \\
    GPT-4.1 & 97.13 & 97.47 & 96.10 & 99.43 & 96.77 & 78.14 \\
    GPT-OSS 120B & 88.51 & 89.87 & 94.16 & 91.43 & 91.61 & 83.61 \\
    Gemini 2.5 P & 95.98 & 94.94 & 94.81 & 99.43 & 96.13 & 89.07 \\
    Gemini 2.5 F & 95.98 & 97.47 & 98.70 & 96.57 & 95.48 & 86.34 \\
    Kimi K2 & 97.13 & 90.51 & 95.45 & 98.86 & 94.84 & 85.79 \\
    DeepSeek v3 & 98.28 & 98.73 & 95.45 & 99.43 & 98.06 & 96.72 \\
    Mistral 24B & 66.67 & 68.35 & 74.68 & 54.86 & 78.06 & 68.85 \\
    Mistral Nemo & 50.57 & 52.53 & 63.64 & 37.14 & 54.19 & 48.09 \\
    Qwen 235B & 97.70 & 98.73 & 99.35 & 99.43 & 98.71 & 87.98 \\
    Llama 70B & 82.76 & 91.77 & 87.66 & 84.57 & 81.29 & 74.86 \\
    Llama 4 & 79.17 & 90.26 & 89.26 & 86.55 & 83.89 & 81.50 \\
    Llama 8B & 31.61 & 30.38 & 35.06 & 29.14 & 32.26 & 26.23 \\
\bottomrule
\end{tabular}
}
\caption{Attribute-level Correctness Rate (CR) for the Retrospective task across six mental state dimensions.}
\label{tab:state_dispersed_results}
\end{table}

\section{Ablations \& Diagnostic Probing}
\label{sec:ablations}
To isolate the mechanisms driving the sharp drop in the Prospective Diagnostic Forecasting performance, we run three targeted diagnostic interventions probing on the main benchmark evaluation set (\S~\ref{sec:empirical_analysis}): (1) model reliance on spurious surface-level shortcuts; (2) robust state-driven reasoning versus static memorization; and (3) structural logic transferability via reasoning injection.

\subsection{Distractor Complexity \& Out-of-Context Exploitation}
\label{subsec:distractor_complexity}
A core structural hypothesis for poor model performance on the Prospective task is its rigorous multiple-choice design: the adversarial distractors share deep semantic boundaries with the ground truth, requiring highly precise social reasoning. We contrast this \emph{Hard Set} against a generated \emph{Easy Set} by replacing adversarial distractors with unrelated out-of-context trajectories sampled randomly across dialogue frames---a distractor generation approach employed in ~\citet{shinoda2025tomato}. This variant introduces distinct topical out-of-distribution shifts, allowing models to identify the correct option purely by leveraging superficial lexical shortcuts or topic coherence without demonstrating genuine functional-ToM reasoning.

As shown in Table~\ref{tab:easy_set_prosp}, lowering the diagnostic barrier boosts model accuracies by 44 to 83 percentage points. Mann-Whitney U tests confirm the Easy--Hard gap is statistically significant across all 13 models and 3 domains ($p < 0.0001$ for 38 of 39 comparisons\footnote{Full statistics in Appendix~\ref{appendix:distractor_complexity}}), with large to very large effect sizes ($r \in [-0.85, -0.40]$). This confirms that the Hard Set successfully eliminates lexical shortcuts, forcing models to rely on state-driven trajectory reasoning alone.

\begin{table}[h]
\centering
\resizebox{0.8\linewidth}{!}{%
\begin{tabular}{lrrr}
\toprule
\multirow{2}{*}{Model} & \multicolumn{3}{c}{\textbf{Prospective-Easy}} \\ \cmidrule{2-4}
 & \multicolumn{1}{c}{MI} & \multicolumn{1}{c}{ESC} & \multicolumn{1}{c}{PFG} \\ \midrule
GPT-5 & 94.85$_{\textcolor{blue}{+80.15}}$ & 91.63$_{\textcolor{blue}{+75.37}}$ & 61.96$_{\textcolor{blue}{+50.54}}$ \\
GPT-4.1 & 91.18$_{\textcolor{blue}{+85.29}}$ & 90.64$_{\textcolor{blue}{+77.83}}$ & 64.13$_{\textcolor{blue}{+58.70}}$ \\
GPT-OSS 120B & 86.76$_{\textcolor{blue}{+77.21}}$ & 78.33$_{\textcolor{blue}{+61.58}}$ & 55.43$_{\textcolor{blue}{+48.91}}$ \\
Gemini 2.5 P & 94.85$_{\textcolor{blue}{+56.62}}$ & 95.07$_{\textcolor{blue}{+66.01}}$ & 67.93$_{\textcolor{blue}{+48.37}}$ \\
Gemini 2.5 F & 92.65$_{\textcolor{blue}{+66.18}}$ & 91.13$_{\textcolor{blue}{+66.01}}$ & 61.41$_{\textcolor{blue}{+51.63}}$ \\
Kimi K2 & 93.38$_{\textcolor{blue}{+71.32}}$ & 89.66$_{\textcolor{blue}{+64.04}}$ & 65.76$_{\textcolor{blue}{+62.50}}$ \\
DeepSeek v3 & 93.38$_{\textcolor{blue}{+83.09}}$ & 87.68$_{\textcolor{blue}{+75.86}}$ & 61.96$_{\textcolor{blue}{+60.33}}$ \\
Mistral 24B & 86.03$_{\textcolor{blue}{+80.15}}$ & 82.76$_{\textcolor{blue}{+77.83}}$ & 65.76$_{\textcolor{blue}{+63.59}}$ \\
Mistral Nemo & 77.21$_{\textcolor{blue}{+70.59}}$ & 66.01$_{\textcolor{blue}{+51.72}}$ & 56.52$_{\textcolor{blue}{+55.43}}$ \\
Qwen 235B & 88.24$_{\textcolor{blue}{+70.59}}$ & 89.66$_{\textcolor{blue}{+60.59}}$ & 57.61$_{\textcolor{blue}{+51.09}}$ \\
Llama 70B & 86.76$_{\textcolor{blue}{+68.34}}$ & 88.18$_{\textcolor{blue}{+68.47}}$ & 58.15$_{\textcolor{blue}{+52.72}}$ \\
Llama 4 & 89.71$_{\textcolor{blue}{+80.15}}$ & 90.64$_{\textcolor{blue}{+78.82}}$ & 69.02$_{\textcolor{blue}{+64.67}}$ \\
Llama 8B & 75.0$_{\textcolor{blue}{+68.38}}$ & 66.5$_{\textcolor{blue}{+49.75}}$ & 50.0$_{\textcolor{blue}{+44.57}}$ \\
\bottomrule
\end{tabular}}
\caption{Performance on the Easy Set of the Prospective task. {\color{blue} Blue} subscripts indicate the performance gains compared directly to the baseline Hard Set.}
\label{tab:easy_set_prosp}
\end{table}

\subsection{Robustness vs. Memorization Verification}
\label{subsec:memorization}
Gemini 3 Pro's uniquely high Prospective performance raises a critical question: does the model demonstrate real Functional-ToM forecasting, or simply exploit benchmark memorization? To test this, we prompt GPT-4o to replace ground-truth mental-state profiles $\mathcal{S}$ with contradictory counterfactual attributes $\mathcal{S}'$ while preserving options (Appendix~\ref{appendix:prompts}), rendering the original ground-truth trajectory logically invalid and forcing a robust reasoner to select ``None of the Above'' (NOTA). We test Gemini 3 Pro alongside a representative cohort of the four strongest and four weakest models.

As summarized in Table~\ref{tab:robustness_memorization}, Gemini 3 Pro maintains high accuracy when identifying NOTA selections under counterfactual conditions ($\text{CR} \in [80.15\%, 95.11\%]$).\footnote{See Appendix~\ref{appendix:false_nota} for a complementary diagnostic.} This resilience disproves data-leak explanations and establishes its robustness under distributional shift. High counterfactual compliance across most baseline architectures---excluding specific model gaps like Llama 8B---confines the core task barrier to robust relational reasoning rather than simple data memorization. Critically, Gemini 3 Pro serves as an empirical proof of concept that deep social-causal trajectory grounding is attainable.

\begin{table}[h]
\centering
\resizebox{0.6\linewidth}{!}{%
\begin{tabular}{llll}
\toprule
\multicolumn{1}{c}{\textbf{Model}} & \textbf{MI} & \textbf{ESC} & \textbf{PFG} \\ \midrule
Gemini 3 P & 80.15 & 94.09 & 95.11 \\
Gemini 2.5 P & 70.59 & 87.68 & 88.59 \\
Gemini 2.5 F & 48.53 & 76.35 & 74.46 \\
GPT-5 & 61.03 & 86.7 & 90.22 \\
Kimi K2 & 80.88 & 95.07 & 95.65 \\ \hdashline
Mistral Nemo & 59.56 & 73.89 & 88.59 \\
Mistral 24B & 97.06 & 99.01 & 99.46 \\
Llama 4 & 67.65 & 78.82 & 55.98 \\
Llama 8B & 37.5 & 37.93 & 51.09 \\
\bottomrule
\end{tabular}%
}
\caption{Robustness vs. Memorization via Counterfactual Intervention. Correctness Rate (CR) for identifying ``None of the Above'' (NOTA).}
\label{tab:robustness_memorization}
\end{table}

\subsection{Teacher-Student Reasoning Injection}
\label{subsec:reasoning_injection}
To test if Gemini 3 Pro's performance stems from its internal logic, we conduct a teacher-student reasoning injection probe targeting weaker student models (Mistral 24B and Mistral Nemo), using GPT-5 as a baseline teacher on the MI subset of Prospective tasks. To ensure that performance shifts are not due to raw label leakage, we processed all traces with Gemini 2.5 Pro zero-shot prompting to mask direct and indirect references to answer options (e.g., ``Option A'') with a \texttt{[MASK]} token prior to injection. The intermediate steps naturally retain the teacher model's chain-of-elimination text, forcing the student model to parse the reasoning by itself.

The results (Table~\ref{tab:reasoning_inject}) reveal a striking performance gap. Injecting Gemini 3 Pro's reasoning traces\footnote{Excerpts are provided in Appendix~\ref{appendix:reasoning_traces_examples}.} triggers a statistically significant performance surge of 42.4-54.1 points over original baseline scores (McNemar's test, $p < 0.01$). In contrast, the negligible 1–5 point shift yielded by GPT-5's traces is statistically indistinguishable from the un-injected baseline. This sharp divergence suggests that Gemini 3 Pro utilizes a highly precise, structured evaluation heuristic that, when externalized, serves as an effective in-context reasoning guide to boost the performance of weaker models\footnote{We further present the qualitative comparison of Gemini 3 Pro's and GPT-5's traces in Appendix~\ref{sectionn:traces_comparison}}.

\begin{table}[h]
\centering
\resizebox{\linewidth}{!}{%
    \begin{tabular}{lccc}
    \toprule
    Model & Baseline & Gemini 3 Pro Trace & GPT-5 Trace \\
    \midrule
    Mistral 24B & 5.88 & 60.0 & 7.0 \\
    Mistral Nemo & 6.62 & 49.0 & 12.0 \\
    \bottomrule
    \end{tabular}
}
\caption{Student model performance when injected with reasoning traces from Gemini 3 Pro vs. GPT-5.}
\label{tab:reasoning_inject}
\end{table}

\section{Conclusion}
We introduced DialToM, a human-verified Theory of Mind (ToM) benchmark designed to evaluate the critical gap between passively inferring mental states and actively utilizing them for behavior prediction in LLMs. Grounded in naturalistic human dialogues and incorporating a novel Trust attribute, DialToM captures the relational dynamics governing human social intelligence. 

Evaluation across 14 LLMs exposed a stark \emph{reasoning asymmetry}. While models demonstrate exceptional proficiency at Retrospective mental state inference ($\ge$80\% accuracy), they collapse on our context-free Prospective Diagnostic Probe, mostly scoring near random chance ($\le$25\%). Rigorous adversarial ablations demonstrated that standard context-driven evaluation tasks allow models to bypass genuine ToM reasoning by exploiting superficial lexical shortcuts and topical coherence. Conversely, Gemini 3 Pro emerged as an exception, establishing a powerful empirical proof of concept that robust, state-driven trajectory grounding is attainable. Furthermore, our Teacher-Student reasoning injection study verified that Gemini 3 Pro's advanced performance stems from an explicit, transferable internal logic that can be successfully externalized in-context to bridge the Functional ToM capability gap in weaker models without requiring parametric updates.

\section{Limitations}
\label{sec:limitations}
Despite the diagnostic rigor of the DialToM framework, several limitations remain.

\paragraph{Trust Dimension Scope} While our introduction of the Trust attribute represents a critical step toward capturing relational rapport, its operational scope is bounded by the specific counselor-client and persuader-persuadee dynamics of our source corpora. Consequently, it may not fully capture broader, cross-cultural, or highly idiosyncratic definitions of interpersonal trust that manifest in casual, non-professional social settings.

\paragraph{Ad-Hoc Trust Grounding} Our pipeline prompts models to generate free-form mental-state inferences, yielding both binary framings (e.g., ``The client views the counselor as trustworthy'') and nuanced relational spectrums (e.g., ``The persuadee views the persuader as moderately trustworthy''), but does not explicitly constrain output generation to established, multi-dimensional psychometric scales (such as standardized benevolence or integrity frameworks). Future benchmark iterations could benefit from grounding Trust generation within structured psychometric scales to ensure greater comparative granularity across items.

\paragraph{Distractor Generation Artifacts} Our data curation pipeline relies on GPT-4o for generating distractors and base mental-state inferences. Although these materials underwent rigorous multi-stage manual filtering via our human-in-the-loop verification framework, we acknowledge the residual risk that subtle, systemic artifacts or linguistic styles may persist in the candidate option distributions.

\paragraph{Restricted Human Baseline} Our crowd worker baseline was structurally restricted to the Retrospective Inference task. Because the context-free Prospective task is cognitively demanding and requires highly specialized domain expertise, it proved too functionally complex to serve as an equitable, unbiased baseline for average, untrained crowd workers. Relatedly, our commitment to high-fidelity adversarial distractors required aggressive Dawid-Skene filtering, leading to lower data retention rates on the Prospective task (e.g., 34\% in the MI domain). While this limits the sheer volume of our benchmark, we accepted this trade-off to guarantee that the final evaluation set is free of semantic ambiguity and rigorously tests Functional-ToM reasoning.

\paragraph{Reliance on LLM-as-a-Judge} While we employ an LLM-as-a-judge protocol for our supplementary semantic alignment analysis (Appendix~\ref{appendix:semantic}), we rely on the established zero-shot evaluative capabilities of frontier models and leave the collection of parallel human meta-judgments for future work. Consequently, these alignment scores are treated as qualitative explorations rather than primary empirical claims.

\paragraph{Passive MCQ Format} DialToM evaluates models using a multiple-choice question (MCQ) framework rather than active generative probing, where models freely generate mental state inferences and dialogue trajectories with higher ecological validity for real-world agents. We chose this structured format for three reasons. First, giving models identical, human-validated mental state profiles establishes a strict baseline. Letting models infer their own initial state risks compound errors, where a downstream failure may simply reflect a poor first guess, whereas a common starting point isolates a model's ability to act on a known state, cleanly decoupling Literal from Functional ToM. 

Second, free-form generation depends on automated metrics (BLEU, ROUGE, BERTScore) or LLM-as-a-Judge protocols, both prone to lexical bias and evaluation artifacts; MCQ instead yields a clean, deterministic Correctness Rate. Third, this format let us evaluate human experts under the same conditions as models. We view MCQ diagnostics and generative probing as complementary, and encourage future work to extend our profiles into open-ended, generative settings.

\paragraph{Temporal and Model Scope} Our findings are bounded by the landscape of proprietary and open-weight models available at testing time, and the severe inference-to-action gaps observed here may shift as scaling laws evolve. Notably, Gemini 3 Pro---the leading baseline in our Prospective Diagnostic evaluation---was discontinued as a preview model on March 9, 2026, and replaced by Gemini 3.1 Pro (Preview)~\cite{googlegemini3discontinue}, so future reproducibility of its results cannot be guaranteed (though Gemini 2.5 Pro remains independently available in our main suite). We have retained all model outputs and will release them upon request for independent verification.

\section{Ethical Considerations}
The curation, annotation, and release of DialToM adhered strictly to established ethical guidelines and were formally approved by the Institutional Review Board (IRB) of Singapore Management University under Approval Number IRB-25-133-A105(725). 

\paragraph{Data Governance and Privacy} DialToM is built entirely from public, open-access datasets (AnnoMI, ESConv, and PersuasionForGood). We complied with all original licenses and user agreements. The source transcripts were fully anonymized by their original authors. We conducted a secondary manual audit to ensure that any remaining names, locations, temporal references, or sensitive behavioral disclosures were completely scrubbed or masked to eliminate re-identification risks.

\paragraph{Crowdsourcing Labor and Compensation} Professional verifiers and crowd workers from Amazon Mechanical Turk (AMT) and Baidu Crowdsourcing were compensated at an average hourly rate of \$12.50 USD, matching or exceeding local living wage standards for digital labor. All participation was strictly voluntary; workers were provided with transparent task descriptions and maintained the right to withdraw at any stage without penalty or loss of accrued compensation.

\paragraph{Dual-Use and Potential Misuse} Advanced Functional ToM models present significant dual-use risks, particularly when deployed in high-stakes helper-recipient environments. In counseling or supportive dialogue  dynamics, failure modes in advanced ToM systems pose direct risks of harm. For example, if a model inaccurately forecasts a user's emotional state or interpersonal trust level, it may generate inappropriate or actively counter-therapeutic advice, potentially exacerbating the user's psychological distress rather than alleviating it.

Moreover, the capacity to track relational trust and accurately forecast state-driven behavior introduces broader social surveillance implications. Models equipped with robust ToM could be exploited by malicious actors to conduct automated, highly personalized behavioral micro-targeting. By precisely mapping and predicting users' latent mental states, these systems could optimize interpersonal manipulation, scale targeted disinformation, and exploit vulnerable demographic groups.

However, concealing evaluation techniques does not prevent these dangers; it simply blinds the AI safety community to potential harm. We release DialToM as a public, non-commercial diagnostic auditing tool to allow researchers to measure, track, and align these social capabilities safely before such systems are deployed in unmonitored, real-world interactions.

\section*{Acknowledgments}
This research is supported by the Singapore Ministry of Health’s National Medical Research Council under its Population Health Research Grant Thematic Category (PHRGTC-5-0005). Any opinions, findings and conclusions or recommendations expressed in this material are those of the author(s) and do not reflect the views of MOH/NMRC. The authors also thank John Pinto for his expert validation of the Prospective tasks, and the anonymous reviewers for their constructive feedback, which greatly strengthened this work.

\bibliography{main}

\appendix

\section{Theoretical Background: Literal vs. Functional Theory of Mind}
\label{appendix:literal_functional_tom}

The distinction between \emph{Literal} and \emph{Functional} Theory of Mind (ToM) addresses a critical methodological flaw in machine ToM evaluations, as formalized by \citet{riemer2025position}. Traditional benchmarks overwhelmingly focus on what the authors define as \emph{Literal Theory of Mind}, which is ``the ability to predict the behavior of others.'' While modern LLMs often achieve high performance on these literal probes, \citet{riemer2025position} caution that such metrics are only informative when models exhibit self-consistent reasoning. In modern LLMs, this consistency cannot be taken for granted. Models can demonstrate strong literal ToM performance that is ultimately illusory, as making accurate predictions about another agent often has ``nothing to do with its own chosen course of behavior.''

To overcome these limitations, \citet{riemer2025position}~introduce the concept of \emph{Functional Theory of Mind}, defined as ``the ability to adapt to agents in-context following a rational response to their behavior.'' Transitioning to a functional framework shifts the benchmarking goal from mere latent state attribution to active forecasting, specifically evaluating ``an agent's ability to choose appropriate actions based on the behavior of another agent.'' DialToM directly operationalizes this theoretical division by structurally partitioning its evaluation workflow into distinct Retrospective and Prospective task suites, enforcing the strict experimental controls necessary to verify that an LLM can actively translate literal inferences into functional behavioral decisions.

\subsection{Methodological Trade-offs of the State-Driven Probing Design}
Our choice to withhold conversational history during the Prospective task introduces an important methodological trade-off regarding task artificiality versus diagnostic rigor.

While \citet{riemer2025position}~conceptualize Functional ToM through interactive, context-heavy multi-agent environments, context-driven forecasting allows models an unwanted degree of freedom. In a multi-turn setup where the history remains visible, LLMs can bypass the complex structural representation of mental states entirely. Instead, the models can achieve high prediction accuracy by relying on local dialogue heuristics, conversational momentum, or superficial lexical matching with the immediate preceding tokens. 

By implementing a \emph{context-free}, \emph{state-driven} configuration, DialToM establishes a strict causal control. Removing the text history breaks the model's reliance on superficial pattern-matching and forces a direct, isolated mapping from the latent mental-state profile to the trajectory choice. If a model successfully completes the task under this setup, it serves as empirically unambiguous proof that the system is treating the explicit mental state primitives as functional, decision-guiding variables rather than passive labels. 

\subsection{Theoretical Basis for Functional ToM Probing}
From a psychometric perspective, validating a cognitive construct requires testing not only whether an internal state can be inferred from observations, but also whether that state systematically guides downstream behavior \cite{cronbach1955construct, whitely1983construct}. Existing ToM benchmarks primarily operationalize the cognitive pipeline in one direction: mapping contextual evidence ($\mathcal{H}$) to latent mental-state representations ($\mathcal{S}$). DialToM complements this paradigm by establishing a strict structural symmetry, isolating the inverse functional mapping from explicit mental-state configurations ($\mathcal{S}$) to downstream behavioral trajectories ($\mathcal{A}$).

This inverse mapping is best evaluated in isolation. Under standard principles of construct representation \cite{whitely1983construct} and nomological validity \cite{cronbach1955construct}, if mapping contextual evidence to a latent state constitutes operational evidence of state representation, then systematically mapping that same state representation to predictable behavioral outcomes constitutes operational evidence of functional state utilization. By hiding conversational history, our framework implements a controlled diagnostic setting standard in cognitive evaluation and psychometric testing (e.g., standardized personality and behavioral questionnaires), where explicit trait or state profiles are evaluated in isolation specifically to determine their predictive impact on downstream actions, independent of surrounding narrative detail. Rejecting this context-free alignment as an invalid abstraction would fundamentally undermine the construct validity of standardized psychometric assessments, which routinely isolate internal state profiles to evaluate behavioral trajectories.

This design pattern also has direct precedent in behavioral psychology. The Theory of Planned Behavior \cite{ajzen1991theory} predicts an individual's action from a compact set of psychological constructs (attitude, subjective norm, perceived behavioral control), without requiring access to the individual's full situational context, precisely because isolating the profile-to-behavior link is what allows a test of whether the profile itself carries predictive information. Our Prospective task applies the same underlying logic to naturalistic dialogue: testing whether an isolated mental-state profile is sufficient to determine a consistent behavioral (dialogue trajectory) outcome.

\subsection{Functional Abstraction and Task Validity} While this design departs from natural human interaction—where predictions are dynamically intertwined with conversational context rather than derived from a static profile---the concrete examples in Appendix~\ref{appendix:task_examples} demonstrate that this abstraction does not undermine construct validity. Across all three conversational domains, candidate trajectories are topically coherent and share deep surface-form similarities. Consequently, they cannot be distinguished via lexical shortcuts or topic coherence. The correct trajectory is identifiable \emph{only} by reasoning about which continuation remains strictly consistent with the speaker's specific mental-state configuration. This design directly operationalizes functional ToM as defined: evaluating the active mapping from explicit mental-state representations to behavioral predictions, rather than surface-level pattern matching against conversational cues.

\subsection{Empirical Validation via Human Baseline} We counter the potential risk of semantic ambiguity or unfair artificiality through our expert validation protocol. If our context-free isolation rendered the mapping inherently ambiguous or textually disjointed, human performance would naturally degrade. However, as shown in Section~\ref{subsec:human_basesline}, our professional domain expert achieved a perfect 100\% score under these exact context-free conditions. This empirical ceiling proves that the isolated mental profiles contain complete, self-contained, and entirely unambiguous causal information capable of guiding correct functional forecasting. Thus, the trade-off favors diagnostic precision without sacrificing the underlying validity of the task.

\section{Detailed Breakdown of Conversational Source Corpus}
\label{sec:appendix_data_sources}

To ground DialToM in naturalistic human interaction, we leverage three distinct, publicly available conversational datasets representing specialized dyadic communication profiles. This section provides the granular background, structural annotations, and linguistic properties of the source material prior to our token normalization and window segmentation pipelines.

\paragraph{AnnoMI} 
Sourced from \citet{annomi2022}, this corpus is uniquely designed to study Motivational Interviewing (MI)---a collaborative, goal-oriented style of communication with particular attention to the language of change, widely utilized in psychotherapy and counseling domains. The raw dataset contains 133 full-length counselor-client video transcripts spanning diverse behavioral topics (e.g., smoking cessation, vaccination hesitancy, and dietary modifications). Every utterance turn is expert-annotated with domain-specific behavioral codes detailing counselor strategies (such as \emph{open questions}, \emph{reflections}, or \emph{persuasion with permission}) and client negotiation postures (\emph{change talk} versus \emph{sustain talk}). We utilize all 133 sessions, treating both high-quality and low-quality execution traces as diverse environmental conditions for Retrospective and Prospective tasks.

\paragraph{ESConv} 
The Emotional Support Conversation corpus \cite{liu-etal-2021-towards} targets the rich cognitive and affective dynamics of empathy and supportive social interaction. It comprises 1,300 multi-turn interaction sessions between human helpers and individuals seeking emotional support. Each interaction is structurally categorized across a taxonomy of help-seeker problems (e.g., academic stress, workplace conflicts, relational distress) and underlying psychological configurations (comprising explicit tags for the seeker's ambient emotion, specific situation, and lived experience). Turns are structurally organized according to progressive emotional support stages: \emph{exploration}, \emph{comforting}, and \emph{action planning}. This progression relies heavily on an agent's ability to recursively model the fluctuating internal distress levels and latent expectations of the speaker.

\paragraph{PersuasionForGood} 
Developed by \citet{wang-etal-2019-persuasion}, this dataset focuses on strategic interpersonal influence and persuasive dialogue. The source corpus captures 1,017 multi-turn natural conversations between human pairs assigned to an asymmetric social task: a \emph{persuader} must convince a \emph{persuadee} to donate a portion of their task compensation to a designated international charity. The data incorporates multi-layered annotations mapping specific communicative strategies used by the persuader (e.g., \emph{logical reasoning}, \emph{emotional appeal}, \emph{credibility framing}) alongside corresponding reactions from the persuadee. Crucially, the dataset provides participant-specific metadata, including self-reported personality profiles (Big Five inventory) and baseline altruism indices, which enrich the cognitive dependencies involved in evaluating strategic interaction trajectories.

\begin{table}[t!]
    \centering
    \resizebox{\linewidth}{!}{
    \begin{tabular}{l rrrr}
    \toprule
         \textbf{Statistic} & \textbf{MI} & \textbf{ESC} & \textbf{PFG} & \textbf{Overall} \\ \midrule
         \multicolumn{5}{c}{\textbf{Original}} \\
         Session Count & 133 & 1300 & 1017 & 2450 \\
         Word Count per session & 1232.97 & 520.61 & 272.78 & - \\
         Turn Count per session & 35.47 & 11.26 & 9.39 \\
         Word Count per turn & 34.76 & 46.22 & 29.07 & - \\ \hline
         \multicolumn{5}{c}{\textbf{Final}} \\
         Context Count & 865 & 3215 & 1863 & 5943 \\
         Word Count per context & 188.2 & 205.02 & 210.8 & - \\
         Turn Count per context & 5.3 & 4.48 & 5.4 & - \\
         Word Count/Turn & 35.5 & 45.5 & 39.1 & - \\
         Retrospective Infer. Qns. & 5190 & 19,290 & 11,178 & 35,658 \\
         Prospective Forecast. Qns. & 865 & 3215 & 1863 & 5943 \\
         Mental State Inference & 100 & 100 & 100  & 300\\
     \bottomrule
    \end{tabular}}
    \caption{Pre- and post-segmentation statistics of the DialToM source corpora. The \textbf{Original} rows report session-level properties of the raw source datasets; the \textbf{Final} rows report context-level properties after sliding window segmentation. Each final context yields 6 Retrospective questions (one per BDIEKT attribute) and 1 Prospective question, totalling $\approx$36,000 evaluation items across the full benchmark suite.}
    \label{tab:full_general_stats}
\end{table}

\section{Detailed Data Annotation Process}
\label{appendix:annotation}

The annotation study was approved by the Institutional Review Board of Singapore Management University (IRB Approval Number: IRB-25-133-A105(725)).

\begin{table*}[t]
\centering
\resizebox{0.8\linewidth}{!}{%
\begin{tabular}{l | rrrr | rrrr | rrrr}
\toprule
\multirow{2}{*}{\textbf{Task}} & \multicolumn{4}{c}{\textbf{MI}} & \multicolumn{4}{c}{\textbf{ESC}} & \multicolumn{4}{c}{\textbf{PFG}} \\ \cmidrule{2-13}
 & O1 & O2 & O3 & O4 & O1 & O2 & O3 & O4 & O1 & O2 & O3 & O4 \\ \midrule
Retrospective & 0.88 & 0.88 & 0.9 & 0.91 & 0.94 & 0.95 & 0.95 & 0.96 & 0.87 & 0.96 & 0.97 & 0.97 \\
Prospective & 0.37 & 0.57 & 0.62 & 0.67 & 0.26 & 0.84 & 0.84 & 0.86 & 0.24 & 0.82 & 0.83 & 0.92 \\
\bottomrule
\end{tabular}}
\caption{Consensus-based verification of task options. Values represent the agreement scores for Professional Verifiers distinguishing the ground-truth (O1) from adversarial distractors (O2–O4).}
\label{tab:task_verification_results}
\end{table*}

\subsection{Crowdsourcing Methodology}

\paragraph{Platform Selection} We chose separate platforms for our tasks by considering worker-task compatibility. To ensure that Lay Participants who author free-form mental state inferences were not exposed to AI-generated inferences before, we recruited Lay Participants for authoring mental state inferences from a different worker pool (AMT) than the verifiers or Lay Participants forming the baseline (Baidu). For the high-volume tasks containing significantly more evaluation samples, we established a contract for mutual benefit with the Baidu Data Crowdsourcing Platform to scale operations efficiently.

\paragraph{Worker Selection} To gather high-quality annotations, we filter workers through a selective worker qualification process. In this process, across both platforms AMT and Baidu, we ask the workers to participate in a replication of the Retrospective task of DialToM. Workers were provided with a conversation excerpt, and are needed to answer a multiple-choice question for selecting the appropriate mental state of the recipient. We selectively chose workers that score at least 80\% on this test.

\subsection{Participant Roles}
We categorized crowd workers into two distinct groups based on the nature of the task:
\begin{enumerate}
    \item \textbf{Professional Verifiers (Baidu):} A cohort of 22 individuals who performed the Verification of AI-generated Mental State Inferences and Dialogue Trajectories. These verifiers were compensated at a competitive hourly rate equivalent to \$12.50/hour, exceeding fair local wage standards. Their high-quality output was aggregated using the Dawid-Skene algorithm to ensure the final dataset's integrity.
    \item \textbf{Lay Participants (Baidu/AMT):} These individuals provided the Human Baseline by completing the benchmarking tasks without prior exposure to the correct labels. Their performance serves to quantify the intrinsic difficulty and domain-specific nuance of DialToM. 
\end{enumerate}

\subsection{Quality Control Methods}
\paragraph{Automated} We employ the Dawid-Skene (DS) algorithm~\cite{Dawid1979} to model annotator skills and verify whether their future annotations remain consistent with previous annotations or not. If the quality diverges significantly, we consider that annotator to be of low-quality and request for a swap of annotations from Baidu.

\paragraph{Manual} We also conduct a high-level quality check manually, by going through all samples and verifying if the annotations are as per what we, as researchers with great deal of experience in this field, would expect.

\subsection{Human Benchmarking \& Expert Ceiling Construction}
\label{subsec:human_benchmarking}
\paragraph{Human Benchmarking} To quantify the intrinsic difficulty of the Retrospective Inference task, three Lay Participants each completed tasks across a stratified subset of 900 evaluation contexts (300 per conversational domain). This data establishes a robust crowd baseline reflecting average human performance. Notably, this crowd benchmarking was restricted exclusively to the Retrospective Inference tasks. This choice was informed by the extreme cognitive load and difficulty of the Prospective Diagnostic Forecasting tasks, which require expert-level social reasoning and long-term logical consistency under a context-free configuration that frequently exceeds the performance bounds of average crowd workers.

\paragraph{Expert Ceiling Validation} To validate the construct validity of the context-free Prospective task and establish a definitive performance ceiling, we recruited a professional licensed counselor specializing in Motivational Interviewing (MI) to complete a stratified sample of 20 Prospective items drawn from the MI domain. These items systematically spanned all difficulty tiers, stratified explicitly by baseline LLM failure rates. A co-author with extensive MI dialogue assessment experience subsequently completed an independent second-pass evaluation blind to both the model outputs and ground-truth labels. Together, their performance establishes a robust expert consensus ceiling.

\subsection{Gold Mental-State Inferences Construction}
Since our gold mental-state inference task is much more complex than other tasks, it thus requires close to native proficiency of speaking, understanding and writing English. Thus, we only chose 9 annotators from AMT that were not only able to reach an optimal score on our qualification test, but those who also held a Master's qualification and were from predominantly English speaking countries in the West, such as United States and the United Kingdom. These annotators provided subjective, free-form mental state inferences for 300 contexts (100 per conversation domain) to evaluate the semantic alignment of AI-generated mental states.

\section{Discussions of Verification Results}
\label{appendix:label_agg}

\begin{table}[h]
\centering
\resizebox{\linewidth}{!}{%
\begin{tabular}{lllllll}
    \toprule
    \multicolumn{1}{c}{\multirow{2}{*}{\textbf{Metric}}} & \multicolumn{3}{c}{\textbf{Retrospective}} & \multicolumn{3}{c}{\textbf{Prospective}} \\ \cmidrule{2-7}
    \multicolumn{1}{c}{} & \multicolumn{1}{c}{\textbf{MI}} & \multicolumn{1}{c}{\textbf{ESC}} & \multicolumn{1}{c}{\textbf{PFG}} & \multicolumn{1}{c}{\textbf{MI}} & \multicolumn{1}{c}{\textbf{ESC}} & \multicolumn{1}{c}{\textbf{PFG}} \\ \midrule
    \# Verification & 400 & 400 & 400 & 400 & 300 & 300 \\
    \# Retained & 306 & 354 & 339 & 136 & 203 & 184 \\
    \% Retention & 76.5 & 88.5 & 84.75 & 34 & 67.67 & 61.33 \\
    Gwet's AC1 & 0.88 & 0.92 & 0.92 & 0.45 & 0.68 & 0.62 \\
    \bottomrule
\end{tabular}
}
\caption{Verification and label aggregation statistics for the Retrospective and Prospective tasks across all three domains. \emph{\# Verification} is the number of samples presented to annotators; \emph{\# Retained} is the 
number passing the Dawid-Skene aggregation threshold; \emph{\% Retention} is the corresponding retention rate; and \emph{Gwet's AC1} measures inter-annotator agreement.}
\label{tab:label_agg}
\end{table}

In Table~\ref{tab:task_verification_results}, we illustrate the average verification statistics averaged over all annotators for both tasks. Following a systematic pattern, all Professional Verifiers found the ``ground truth'' (O1) much harder (lower score here implies confusion and increased difficulty of the task) to differentiate from the other options.

Table~\ref{tab:label_agg} shows the dataset statistics post human verification and Dawid-Skene label aggregation. We were able to retain $\approx$ 80\% of the data for the Retrospective and $\approx$ 60\% for the Prospective task.

Manual error auditing indicates that the lower consensus and data retention observed in the Prospective task is a direct structural consequence of our rigorous diagnostic probing design. The adversarial distractors were deliberately engineered to share high surface-level text and topic similarities with the ground-truth continuation. While this intentionally sets a high-difficulty threshold that challenges average professional verifiers and crowd workers, our expert ceiling check confirms that the items remain completely unambiguous to domain specialists.

\begin{table*}[t]
\centering
\resizebox{0.6\linewidth}{!}{%
\begin{tabular}{lrrr}
\toprule
    Model & Parameter Size & Developer & Knowledge Cut-off \\ \midrule
    \texttt{Mistral Nemo} & 12B & Mistral AI & April 2024 \\
    \texttt{Mistral 3.2 24B} & 24B & Mistral AI & October 2023 \\
    \texttt{Moonshot Kimi K2} & 1T, 32B active & Moonshot AI & $\sim$October 2024 \\
    \texttt{Qwen3 235B} & 235B, 22B active & Alibaba & June 2025 \\
    \texttt{Llama 4 Maverick} & 400B, 17B active & Meta AI & August 2024 \\
    \texttt{Llama 3.3 70B} & 70B & Meta AI & December 2023 \\
    \texttt{Llama 3.1 8B} & 8B & Meta AI & December 2023 \\
    \texttt{DeepSeek v3} & 671B, 37B active & DeepSeek & March 2025 \\ \hline
    \texttt{Gemini 3 Pro}$^\dagger$ & - & Google & - \\
    \texttt{Gemini 2.5 Flash} & - & Google & - \\
    \texttt{Gemini 2.5 Pro} & - & Google & - \\
    \texttt{GPT-OSS 120B} & 120B & OpenAI & June 2024 \\
    \texttt{GPT-4.1} & - & OpenAI & - \\
    \texttt{GPT-5} & - & OpenAI & - \\
    \bottomrule
\end{tabular}}
\caption{Overview of evaluated LLMs, categorized by architecture and access type. Technical specifications and knowledge cut-offs are provided where publicly available. Models above the horizontal rule are open-weights; models below are closed-source. $^\dagger$\texttt{gemini-3-pro-preview} was discontinued by Google on March 9, 2026; see \S~\ref{appendix:models} for details.}
\label{tab:model_list}
\end{table*}

\begin{table*}[ht]
\centering
\resizebox{0.8\linewidth}{!}{%
\begin{tabular}{lccc}
\hline
\textbf{Parameter} & \textbf{OpenAI} & \textbf{Google} & \textbf{OpenRouter} \\
\hline
Temperature & 1.0 & Model-specific default & 1.0 \\
Top-$p$ & 1.0 & 0.95 & 1.0 \\
Top-$k$ & N/A & 40 & 0 \\
Max output tokens & Model context length & Model context length (65,535) & Model context length \\
Presence penalty & 0.0 & 0.0 & 0.0 \\
Frequency penalty & 0.0 & 0.0 & 0.0 \\
Seed & None (random) & None (random) & None (random) \\
Stop sequences & None & None & None \\
\hline
\end{tabular}}
\caption{Default generation parameters across LLM providers.}
\label{tab:default_api_parameters}
\end{table*}

\section{Evaluated Models}
\label{appendix:models}

Table~\ref{tab:model_list} summarizes the 14 LLMs evaluated in this study. They span parameter scales from 12B to 1T, covering both standard dense and Mixture-of-Experts (MoE) architectures. Closed-source proprietary models (Gemini family, GPT family) are queried via official provider APIs; open-weights models are queried via the OpenRouter API. Parameter sizes and knowledge cut-offs are reported where publicly disclosed by the respective developers; entries marked ``-'' reflect undisclosed specifications.

To ensure full reproducibility, Table~\ref{tab:default_api_parameters} reports the exact default generation parameters used for every provider, applied uniformly across all evaluated models regardless of whether native reasoning was permitted during inference (\S\ref{sec:experiments}). We use each provider's default parameters uniformly to avoid granting any model a decoding-level advantage; for reasoning models specifically, this also follows provider guidance, as Google's Gemini 3 API documentation recommends against lowering temperature from its default of 1.0, warning that doing so may cause looping or degraded performance on complex reasoning tasks.

\paragraph{Note on Gemini 3 Pro Availability} Gemini 3 Pro (\texttt{gemini-3-pro-preview}) was available via the Google Gemini API in preview mode at the time of our study. However, Google officially discontinued \texttt{gemini-3-pro-preview} on March 9, 2026, replacing it  with Gemini 3.1 Pro Preview (\texttt{gemini-3.1-pro-preview})~\cite{googlegemini3discontinue}. As a result, direct replication of our Gemini 3 Pro results using the current Google API is no longer possible. Researchers wishing to verify our findings should note that Gemini 2.5 Pro is independently included in the main benchmarking results (Table~\ref{tab:benchmarking_results}) and serves as the representative Gemini-family model in the Dialogue Context Degradation Analysis (\S~\ref{appendix:context_degradation}). All raw model outputs from our original Gemini 3 Pro benchmarking runs have been preserved and are available upon request.

\section{Dialogue Context Degradation Analysis}
\label{appendix:context_degradation}

To ground DialToM's context-free, state-driven design within a broader evaluation continuum and directly contrast it with context-present frameworks like SimpleToM~\cite{gu2024simpletom} and RecToM~\cite{li2025rectombenchmarkevaluatingmachine}, we systematically vary the availability of conversational context. We evaluate these structural configurations across a stratified diagnostic sample of 100 evaluation items (34 MI, 33 ESC, and 33 PFG), maintaining a completely paired experimental design across three representative models: Gemini 2.5 Pro, GPT-5, and Mistral 24B.

We define three specific experimental conditions representing progressive levels of contextual exposure:
\begin{itemize}[noitemsep]
    \item \textbf{State-Only}: The canonical DialToM setup. Models are provided exclusively with the latent mental state profile in isolation, establishing the most stringent diagnostic pressure by eliminating all conversational clues.
    \item \textbf{Summary + States}: A dialogue summary generated by Gemini 2.5 Pro\footnote{See Appendix~\ref{appendix:prompts} for the prompt template.} is prepended to the core mental state profile, introducing coarse-grained topical contexts while withholding fine-grained lexical patterns.
    \item \textbf{Conversation History + States}: The complete multi-turn conversation history is fully retained and prepended to the profile, closely mirroring conventional context-present evaluation configurations (e.g.,~\cite{gu2024simpletom,li2025rectombenchmarkevaluatingmachine}).
\end{itemize}

\begin{figure*}[t]
    \centering
    \includegraphics[width=\linewidth]{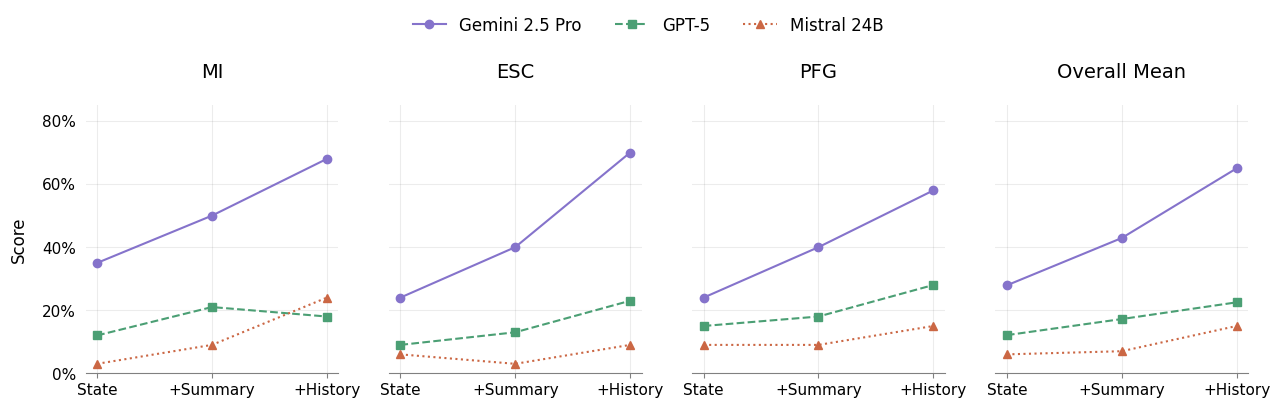}
    \caption{Performance gradient across context conditions}
    \label{fig:ctx_degrad}
\end{figure*}

Results are shown in Figure~\ref{fig:ctx_degrad}. A clear, monotonic performance gradient emerges across all three conditions, with accuracy increasing as contextual exposure scales up. The effect is pronounced for Gemini 2.5 Pro (27.9\% $\rightarrow$ 42.9\% $\rightarrow$ 65.0\%), confirming that conversational context provides meaningful signal and that its absence imposes significant performance pressure. However, GPT-5 and Mistral 24B remain near or below the 25\% random baseline even under full-dialogue history conditions (22.5\% and 15.8\%, respectively), demonstrating that their failure on the Prospective task cannot be attributed to the absence of dialogue history alone. These models lack the underlying functional-ToM reasoning capacity regardless of contextual scaffolding. Together, these findings validate DialToM's State-Driven Diagnostic Probe---the context-free condition maximizes diagnostic pressure while remaining objectively solvable for domain experts (\S~\ref{subsec:human_basesline}), isolating true functional-ToM reasoning from surface-level context exploitation.

\section{Semantic Comparison with Gold Reference}
\label{appendix:semantic}

\paragraph{NLP Metrics}
To compare LLM-generated and human-written mental-state inferences, we utilize standard semantic similarity metrics commonly used in NLP evaluations. We adopt both linguistic metrics (\textbf{BLEU} and \textbf{ROUGE-L}) to measure the linguistic similarity between two texts, and semantic similarities (\textbf{BERTScore}) which compare between the generated text of the LLM inference and the gold reference by measuring the cosine similarity between embeddings of two pieces of text. We report the average similarity scores across all references to handle multiple gold references.

In Table~\ref{tab:written_inference}, we compare the results of LLM-generated inferences with that of human-written gold references. The first row, GPT-4o, shows the similarities between the inferences used in the benchmarking tasks and human annotation.

We find LLMs are only moderately similar to human written gold inferences by semantic similarity (signified by moderate BERTScores) and very low overlap measured by linguistic similarity (signified by low BLEU, ROUGE-L scores). These overall low scores suggest the challenging aspect of generating free-form mental state inferences and the existing lexical diversity in generations while preserving semantic. Nevertheless, the proprietary LLMs (GPT and Gemini families) perform better relative to other models on writing tasks. This overall weak semantic similarity creates a dissonance with respect to the otherwise strong Retrospective task performance.

\begin{table}[h]
\centering
\resizebox{0.8\linewidth}{!}{%
\begin{tabular}{lrrr}
\toprule
    \multicolumn{1}{c}{\multirow{2}{*}{\textbf{Model}}} & \multicolumn{3}{c}{\textbf{Gold Mental-State Reference}} \\ \cmidrule{2-4} 
    \multicolumn{1}{c}{} & \multicolumn{1}{c}{BLEU} & \multicolumn{1}{c}{ROUGE-L} & \multicolumn{1}{c}{BERTScore} \\ \midrule
    GPT-4o & 9.44 & 0.31 & 0.44 \\ \hdashline
    GPT-5 & 8.56 & 0.32 & 0.41 \\
    GPT-4.1 & 9.71 & 0.33 & 0.43 \\
    GPT-OSS 120B & 8.09 & 0.32 & 0.41 \\
    Gemini 2.5 P & 8.90 & 0.32 & 0.44 \\
    Gemini 2.5 F & 7.98 & 0.32 & 0.39 \\
    Kimi K2 & 7.77 & 0.29 & 0.38 \\
    DeepSeek v3 & 8.75 & 0.32 & 0.42 \\
    Mistral 24B & 6.98 & 0.29 & 0.37 \\
    Mistral Nemo & 6.21 & 0.29 & 0.36 \\
    Qwen 235B & 8.05 & 0.30 & 0.40 \\
    Llama 70B & 6.67 & 0.28 & 0.34 \\
    Llama 4 & 7.32 & 0.29 & 0.35 \\
    Llama 8B & 5.30 & 0.25 & 0.31 \\
\bottomrule
\end{tabular}
}
\caption{Linguistic alignment between AI and human-written inferences. Low linguistic overlap (BLEU/ROUGE-L) alongside moderate BERTScore values indicate high lexical diversity despite semantic preservation.}
\label{tab:written_inference}
\end{table}

\paragraph{LLM-as-a-Judge Evaluation}
To complement the standard NLP metrics above, we conducted an exploratory LLM-as-a-Judge evaluation\footnote{The evaluation prompt is provided in Appendix~\ref{appendix:prompts}.} using Gemini 3 Pro\footnote{Gemini 3 Pro was discontinued on March 9, 2026 (Appendix~\ref{appendix:models}); all judge outputs are preserved and available upon request.} as a zero-shot judge on LLM-generated inferences to measure how their mental-state inferences align with our human gold references. The LLM judge was provided with the original source dialogue alongside the human and AI inferences, ensuring it evaluated alignment based on the conversational context, rather than lexical overlap. The judge compared these samples across three dimensions on a 3-point Likert scale: 
\begin{itemize}
\item \textbf{Propositional Agreement} (PA): Do the AI and human inferences express the same core proposition about the mental state?
\item \textbf{Causal Attribution} (CA): Do the AI and human anchor the mental state to the same conversational cause or evidence? 
\item \textbf{Specificity Alignment} (SA): Are the AI and human inferences operating at the same level of specificity/granularity?
\end{itemize}

\begin{table}[h]
\centering
\resizebox{0.85\linewidth}{!}{%
\begin{tabular}{lrrrr}
\toprule
    \multicolumn{1}{c}{\multirow{2}{*}{\textbf{Model}}} & \multicolumn{4}{c}{\textbf{Gold Mental-State Reference}} \\ \cmidrule{2-5}
    & PA ($/3$) & CA ($/3$) & SA ($/3$) & TA ($/9$) \\ \midrule
    GPT-5 & 1.78 & 2.03 & 1.53 & 5.33 \\
    GPT-4.1 & 1.73 & 2.03 & 1.61 & 5.37 \\
    GPT-OSS 120B & 1.65 & 1.82 & 1.45 & 4.91 \\
    Gemini 2.5 P & 1.66 & 1.99 & 1.55 & 5.19 \\
    Gemini 2.5 F & 1.53 & 1.74 & 1.37 & 4.65 \\
    Kimi K2 & 1.58 & 1.94 & 1.45 & 4.97 \\
    DeepSeek v3 & 1.72 & 2.07 & 1.60 & 5.39 \\
    Mistral 24B & 1.60 & 1.97 & 1.48 & 5.05 \\
    Mistral Nemo & 1.25 & 1.40 & 1.13 & 3.79 \\
    Qwen 235B & \textbf{1.82} & \textbf{2.17} & \textbf{1.65} & \textbf{5.63} \\
    Llama 70B & 1.62 & 1.94 & 1.48 & 5.04 \\
    Llama 4 & 1.51 & 1.84 & 1.42 & 4.77 \\
    Llama 8B & 1.16 & 1.52 & 1.19 & 3.87 \\
\bottomrule
\end{tabular}
}
\caption{Dialogue-anchored semantic comparison using an LLM-as-a-Judge approach to directly measure causal alignment between written inferences. PA: Propositional Agreement; CA: Causal Attribution Agreement; SA: Specificity Alignment; TA: Total Alignment Score. PA, CA, SA are all scored on a Likert scale of 3, whereas TA = PA + CA + SA, and is scored out of 9.}
\label{tab:llm_judge_written}
\end{table}

Table~\ref{tab:llm_judge_written} outlines the alignment statistics across all evaluated models (excluding Gemini 3 Pro as the evaluator). Qwen 235B yields the closest alignment to human references across all axes, peaking at a Causal Attribution (CA) score of 2.17/3. On average, models consistently display high Causal Attribution (CA), followed by moderate Propositional Agreement (PA), while exhibiting noticeable divergence on Specificity Alignment (SA). 

This uniform pattern indicates that while LLMs successfully isolate and anchor their reasoning to identical conversational triggers as humans, they tend to generate coarse-grained semantic generalizations. In contrast, human annotators formulate highly individualistic, context-specific assertions. These alignment variances indicate that the low SA trends reflect structural differences in abstraction preferences rather than flaws in logical grounding, providing qualitative evidence of how AI-generated mental-state reasoning diverges from human baseline.

\section{Distinguishing Genuine Reasoning from Confusion in NOTA Selection}
\label{appendix:false_nota}

Table~\ref{tab:robustness_memorization} (\S\ref{subsec:memorization}) reports accuracy exclusively on tasks where ``None of the Above'' (NOTA) is the correct answer. This design cannot, on its own, distinguish genuine counterfactual robustness from a model that systematically defaults to NOTA whenever it is confused, which would artificially inflate scores for weaker models without reflecting real relational reasoning.

To directly examine this, we construct a targeted false-NOTA diagnostic. For a random sample of 100 items per domain (300 total across MI, ESC, PFG), we take the original main Prospective task items (unmodified mental-state profiles, correct trajectories, and distractors) and add NOTA as a 5th option. This lets us measure the \textit{false-NOTA error rate}: how often a model incorrectly selects NOTA when a genuinely correct answer is available among the options, paired against \textit{true-NOTA accuracy} computed on a matched 100-item-per-domain subsample drawn from the original counterfactual pool.\footnote{True-NOTA figures here are computed on this matched subsample and may differ slightly from the full-set results reported in Table~\ref{tab:robustness_memorization}.} We evaluate GPT-5, Gemini 2.5 Pro, Gemini 3.1 Pro, Mistral 24B, and Mistral Nemo, using Gemini 3.1 Pro as a replacement for the now-discontinued Gemini 3 Pro (\S\ref{sec:limitations}).

\begin{table}[h]
\centering
\resizebox{\linewidth}{!}{
\small
\begin{tabular}{lcccccc}
\toprule
& \multicolumn{3}{c}{True-NOTA (\%)} & \multicolumn{3}{c}{False-NOTA FPR (\%)} \\
Model & MI & ESC & PFG & MI & ESC & PFG \\
\midrule
Gemini 3.1 Pro & 93.0 & 99.0 & 93.0 & 4.0 & 3.6 & 0.0 \\
GPT-5 & 62.0 & 79.0 & 87.0 & 0.0 & 2.0 & 1.0 \\
Gemini 2.5 Pro & 53.0 & 72.0 & 76.0 & 0.0 & 0.0 & 0.0 \\
Mistral 24B & 81.0 & 86.0 & 98.0 & 0.0 & 0.0 & 1.0 \\
Mistral Nemo & 73.0 & 76.0 & 86.0 & 9.0 & 11.0 & 4.0 \\
\bottomrule
\end{tabular}
}
\caption{True-NOTA accuracy vs. False-NOTA false positive rate (FPR), on matched 100-item-per-domain samples.}
\label{tab:false_nota}
\end{table}

Across all five models, false-NOTA false positive rates remain low (0--11\%), with the highest rates observed for the weakest-performing model on the main Prospective task (Mistral Nemo, 4--11\%). This is far below what would be expected if confusion-driven NOTA-defaulting explained the high true-NOTA accuracy observed across all models (53--99\%). If weaker models were systematically defaulting to NOTA under confusion, false-NOTA error rates should approach the same magnitude as true-NOTA accuracy, rather than remaining an order of magnitude lower. Notably, even Mistral Nemo's true-NOTA accuracy (73--86\%) vastly exceeds its false-NOTA error rate (4--11\%), indicating its NOTA selections follow real profile-contradiction signals rather than an indiscriminate NOTA preference.

We attribute the low false-NOTA rates to our adversarial distractor design. Distractors are deliberately constructed to be maximally topically and lexically similar to the correct trajectory, making them categorically more salient to a model exploiting surface coherence than an abstract NOTA option. A confused model is therefore more likely to gravitate toward a specific, plausible-looking distractor than to default to NOTA. This pattern holds across all five evaluated models, indicating the NOTA methodology is sound and is not an artifact of any single model's behavior.

\section{Qualitative Trace Comparison} 
\label{sectionn:traces_comparison}

To investigate what qualtitatively distinguishes Gemini 3 Pro's reasoning from GPT-5's (\S\ref{subsec:reasoning_injection}), we conduct a blind pairwise evaluation of 50 masked reasoning trace pairs randomly sampled from the Prospective tasks' motivational interviewing (MI) subset. Each pair consists of one Gemini 3 Pro trace and one GPT-5 trace, assigned randomly as traces A and B in the evaluation. Claude Opus 4.8 was utilized as the underlying judge model to score each trace along four dimensions on a 0--2 scale\footnote{Full judge prompt in Appendix~\ref{appendix:prompts}.}:

\begin{itemize}
\item \textbf{Attribute Coverage}: the number of mental-state attributes (BDIEKT) the trace explicitly engages.
\item \textbf{Elimination Specificity}: the degree to which rejected options are tied to specific attribute-level conflicts.
\item \textbf{Consistency Verification}: the extent to which the selected option is checked against the full mental-state profile before concluding.
\item \textbf{Surface vs.\ Structural Language}: the extent to which the reasoning relies on topical similarity vs. logical consistency with stated attributes.
\end{itemize}

As shown in Table~\ref{tab:qualitative_judge}, Gemini 3 Pro scores significantly higher than GPT-5 on all four dimensions (Wilcoxon signed-rank test, $p < 0.01$ per dimension; $p < 0.0001$ on total score). The gap is largest on \textit{Surface vs.\ Structural Language} and \textit{Consistency Verification}, suggesting that Gemini's traces are distinguished less by covering more attributes per se than by explicitly verifying the selected option against the full profile and grounding rejections in specific attribute conflicts rather than topical resemblance. Interestingly, five GPT-5 traces did not engage with the mental-state profile at all, but instead reasoning about unrelated content.

\begin{table}[h]
\centering
\resizebox{0.9\linewidth}{!}{
\begin{tabular}{lcc}
\toprule
Dimension & Gemini 3 Pro & GPT-5 \\
\midrule
Attribute Coverage & 1.92 & 1.42 \\
Elimination Specificity & 1.04 & 0.64 \\
Consistency Verification & 1.44 & 0.86 \\
Surface vs.\ Structural & 1.62 & 0.96 \\
\midrule
Total (ouf of 8) & 6.02 & 3.88 \\
\bottomrule
\end{tabular}}
\caption{Blind pairwise judge scores on 50 masked Prospective-MI trace pairs. All per-dimension and total differences are significant under paired Wilcoxon tests ($p < 0.01$).}
\label{tab:qualitative_judge}
\end{table}

Table~\ref{tab:trace_excerpt} displays a representative pair illustrating the qualitative distinction. Gemini's trace evaluates each masked option against specific mental-state attributes before converging on a match. In contrast, GPT-5's trace identifies the same general theme but stops after a single surface-level pass without checking the selection against the remaining attributes.

\begin{table}[h]
\centering
\small
\begin{tabular}{p{7.2cm}}
\toprule
\textbf{Gemini 3 Pro (Attribute Coverage: 2, Consistency Verification: 2)} \\
``[MASK] touches on peer pressure, which is relevant, but the client's response feels generic. It addresses the issue without specifically honing in on the desire to fit in. [...] Now, [MASK]. The counselor suggests reasons for drinking: [...] \textit{fear of being disliked by friends}. Ding, ding, ding! This directly hits the client's core vulnerability [...] The client's response [...] confirms the `hard to avoid because of social situations' belief.'' \\
\midrule
\textbf{GPT-5 (Attribute Coverage: 1, Consistency Verification: 0)} \\
``The most plausible direction is where the client opens up about peer pressure. Option [MASK] directly addresses this, emphasizing the struggle to fit in, which resonates with the client's feelings and group norms.'' \\
\bottomrule
\end{tabular}
\caption{Excerpted, masked reasoning traces for the same Prospective-MI item. Gemini checks the candidate against multiple stated attributes (belief, desire, and emotion) before selecting; GPT-5 identifies the same surface theme but does not verify against the remaining profile attributes.}
\label{tab:trace_excerpt}
\end{table}

\section{Statistical Analysis}
\label{appendix:statistical_analysis}

\subsection{Inference-to-Action Gap}
\label{appendix:inference-to-action_gap}
Table~\ref{tab:chisquare_full} reports the full Pearson's Chi-Square statistics ($\chi^2$), $p$-values, and Cram\'er's $V$ effect sizes for the Retrospective--Prospective performance gap across all 14 models and 3 domains. All 42 comparisons are statistically significant at $\alpha = 0.01$, and 41 of 42 at $\alpha = 0.001$, confirming that the inference-to-action gap is a robust, systematic phenomenon rather than an artifact of any particular model or domain.

Effect sizes span a wide range ($V \in [0.11, 0.90]$), reflecting meaningful variation in the magnitude of the gap across models. Gemini 3 Pro yields the smallest effect sizes ($V \in [0.24, 0.33]$), consistent with its uniquely narrow Retrospective--Prospective gap and its strong Functional ToM capability. At the opposite extreme, DeepSeek V3 exhibits the largest effect sizes ($V \in [0.87, 0.90]$), indicating the most pronounced asymmetry of any evaluated model. The single weakest result is Llama 3.1 8B on ESC ($\chi^2 = 7.19$, $p = 0.007$, $V = 0.114$); notably, this small effect does not reflect competence on the Prospective task but rather near-chance performance on both tasks, making the gap itself narrow by floor effect rather than by functional-ToM ability.

Effect sizes are generally consistent across domains within each model, with no systematic domain ordering, suggesting that the inference-to-action gap is a model-level property rather than a domain-specific artifact.

\begin{table*}[t]
\centering
\resizebox{0.8\linewidth}{!}{
\begin{tabular}{l rrr rrr rrr}
\toprule
\multirow{2}{*}{\textbf{Model}} 
    & \multicolumn{3}{c}{\textbf{MI}} 
    & \multicolumn{3}{c}{\textbf{ESC}} 
    & \multicolumn{3}{c}{\textbf{PFG}} \\
\cmidrule(lr){2-4} \cmidrule(lr){5-7} \cmidrule(lr){8-10}
    & \textbf{$\chi^2$} & \textbf{$p$} & \textbf{$V$}
    & \textbf{$\chi^2$} & \textbf{$p$} & \textbf{$V$}
    & \textbf{$\chi^2$} & \textbf{$p$} & \textbf{$V$} \\
\midrule
Gemini 3 Pro     & 29.56  & $<$0.001 & 0.259 
                 & 61.78  & $<$0.001 & 0.333 
                 & 30.83  & $<$0.001 & 0.243 \\
Gemini 2.5 Pro   & 148.08 & $<$0.001 & 0.579 
                 & 295.81 & $<$0.001 & 0.729 
                 & 310.66 & $<$0.001 & 0.771 \\
Gemini 2.5 Flash & 223.40 & $<$0.001 & 0.711 
                 & 299.86 & $<$0.001 & 0.734 
                 & 314.34 & $<$0.001 & 0.775 \\
GPT-5            & 270.48 & $<$0.001 & 0.782 
                 & 356.98 & $<$0.001 & 0.801 
                 & 331.87 & $<$0.001 & 0.797 \\
GPT-4.1          & 316.89 & $<$0.001 & 0.847 
                 & 368.29 & $<$0.001 & 0.813 
                 & 362.13 & $<$0.001 & 0.832 \\
GPT-OSS 120B     & 173.56 & $<$0.001 & 0.627 
                 & 291.37 & $<$0.001 & 0.723 
                 & 299.39 & $<$0.001 & 0.757 \\
Kimi K2          & 219.48 & $<$0.001 & 0.705 
                 & 296.66 & $<$0.001 & 0.730 
                 & 333.49 & $<$0.001 & 0.799 \\
Qwen3 235B       & 272.05 & $<$0.001 & 0.785 
                 & 295.81 & $<$0.001 & 0.729 
                 & 358.25 & $<$0.001 & 0.828 \\
DeepSeek V3      & 331.74 & $<$0.001 & 0.866 
                 & 423.20 & $<$0.001 & 0.872 
                 & 422.99 & $<$0.001 & 0.899 \\
Mistral 3.2 24B  & 114.26 & $<$0.001 & 0.508 
                 & 208.73 & $<$0.001 & 0.612 
                 & 209.82 & $<$0.001 & 0.633 \\
Mistral Nemo     & 53.74  & $<$0.001 & 0.349 
                 & 77.04  & $<$0.001 & 0.372 
                 & 129.75 & $<$0.001 & 0.498 \\
Llama 4 Maverick & 244.15 & $<$0.001 & 0.743 
                 & 305.20 & $<$0.001 & 0.740 
                 & 258.86 & $<$0.001 & 0.704 \\
Llama 3.3 70B    & 133.65 & $<$0.001 & 0.550 
                 & 226.78 & $<$0.001 & 0.638 
                 & 320.01 & $<$0.001 & 0.782 \\
Llama 3.1 8B     & 28.66  & $<$0.001 & 0.255 
                 & 7.19   & 0.007    & 0.114$^\dagger$
                 & 35.01  & $<$0.001 & 0.259 \\
\bottomrule
\end{tabular}}
\caption{Pearson's Chi-Square statistics ($\chi^2$), $p$-values, and Cram\'er's $V$ effect sizes for the Retrospective vs.\ Prospective performance gap, per model and domain. All comparisons are significant at $\alpha = 0.01$. $^\dagger$Llama 3.1 8B on ESC returns the smallest effect ($V = 0.114$), indicating near-chance performance on both tasks.}
\label{tab:chisquare_full}
\end{table*}

\subsection{Error Analysis of the Lay Participant Baseline}
\label{subsec:human_error_analysis}

The low Retrospective lay baseline (\S\ref{subsec:human_basesline}) raises a natural question whether this reflect ambiguous instructions or genuine task difficulty. Therefore, we conduct a two-part error analysis on all 900 Retrospective items completed by three Lay Participants each (2,700 total choices across MI, ESC, PFG).

\paragraph{Positional Bias} We first test if Lay Participants favor certain screen positions (A/B/C/D) regardless of content. A chi-square goodness-of-fit test indicates a modest but significant positional bias in MI and ESC, but not PFG (Table~\ref{tab:position_bias}); pooled across all domains, the effect remains weakly significant ($\chi^2 = 9.58$, $p = 0.0225$).\footnote{Choices from the same worker across items are correlated, so the effective $N$ is closer to the number of workers than the total choice count; the reported $p$-values are therefore anti-conservative and should be read as an upper bound on significance.} Effect sizes are modest overall and do not account for the magnitude of the low lay baseline on their own.

\begin{table}[h]
\centering
\resizebox{0.9\linewidth}{!}{
\begin{tabular}{lccccc}
\toprule
Domain & A & B & C & D & $\chi^2$ ($p$) \\
\midrule
MI & 258 & 194 & 206 & 242 & 12.00 (0.0074) \\
ESC & 216 & 270 & 193 & 221 & 13.98 (0.0029) \\
PFG & 253 & 215 & 215 & 217 & 4.66 (0.1986) \\
\midrule
Pooled & 727 & 679 & 614 & 680 & 9.58 (0.0225) \\
\bottomrule
\end{tabular}}
\caption{Lay Participant choice counts by screen position (A/B/C/D), with chi-square goodness-of-fit test against a uniform null.}
\label{tab:position_bias}
\end{table}

\paragraph{Distractor Bias} Next, we test whether incorrect responses cluster on a single, specific distractor per item, or spread uniformly across all three distractors. For each item, we calculate the normalized Shannon entropy $H_n$ over the three distractor counts among incorrect responses ($H_n = 1$ indicates a uniform split across the three distractors; $H_n = 0$ indicates total agreement on one distractor), then take the weighted mean across items (weighted by number of incorrect responses per item, restricted to items with $\geq 2$ incorrect responses to yield any distributional information).

The results reveal extreme clustering (Table~\ref{tab:distractor_bias}). The pooled weighted mean is $H_n = 0.083$ across 1,897 incorrect responses covering 657 items. MI ($H_n = 0.094$) and ESC ($H_n = 0.033$) show the most severe clustering, whereas, PFG ($H_n = 0.125$) is modestly higher but still far from uniform.

\begin{table}[h]
\centering
\resizebox{\linewidth}{!}{
\begin{tabular}{lccc}
\toprule
Domain & Items ($\geq$2 incorrect) & Incorrect Responses & $H_n$ \\
\midrule
MI & 206 / 300 & 590 & 0.094 \\
ESC & 228 / 300 & 669 & 0.033 \\
PFG & 223 / 300 & 638 & 0.125 \\
\midrule
Pooled & 657 / 900 & 1{,}897 & 0.083 \\
\bottomrule
\end{tabular}}
\caption{Weighted mean normalized entropy ($H_n$) over incorrect Lay Participant responses. Lower values indicate stronger clustering on a single distractor.}
\label{tab:distractor_bias}
\end{table}

Overall, these results indicate that when Lay Participants answered incorrectly, they were not guessing randomly across all options, but were generally drawn toward the same specific distractor. This is the expected behavior of our adversarial distractor design, which deliberately makes distractors share deep semantic and surface-level similarity with the ground-truth answer. Near-total distractor agreement, rather than low-confidence guessing, suggests that the low lay baseline reflects the intended difficulty of a well-specified but adversarially challenging task, instead of unclear instructions or poor task calibration.

\subsection{Distractor Complexity}
\label{appendix:distractor_complexity}
Table~\ref{tab:mannwhitney_full} reports the full Mann-Whitney U statistics and rank-biserial effect sizes ($r$) for the Easy--Hard distractor comparison across all 13 models and 3 domains. All 39 comparisons are statistically significant ($\alpha = 0.05$). Effect sizes are large to very large throughout ($|r| \geq 0.39$ in all 39 cases, and $\geq 0.40$ in 38 cases), with stronger effects in MI and ESC ($r \in [-0.853, -0.566]$ and $r \in [-0.788, -0.498]$, respectively) than in PFG ($r \in [-0.587, -0.397]$), indicating that persuasion-domain dialogue trajectories are inherently more topically distinct, modestly reducing the diagnostic pressure of adversarial distractors in that domain. The single weakest result in terms of statistical significance is Gemini 2.5 Pro on PFG ($p = 0.012$, $r = -0.429$), though it easily clears the $\alpha = 0.05$ threshold and still represents a large effect.

\begin{table*}[t]
\centering
\resizebox{0.7\linewidth}{!}{
\begin{tabular}{l rr rr rr}
\toprule
\multirow{2}{*}{\textbf{Model}} 
    & \multicolumn{2}{c}{\textbf{MI}} 
    & \multicolumn{2}{c}{\textbf{ESC}} 
    & \multicolumn{2}{c}{\textbf{PFG}} \\
\cmidrule(lr){2-3} \cmidrule(lr){4-5} \cmidrule(lr){6-7}
    & \textbf{U} & \textbf{$r$} 
    & \textbf{U} & \textbf{$r$} 
    & \textbf{U} & \textbf{$r$} \\
\midrule
GPT-4.1         & 17136.0 & $-0.853$ & 36641.5 & $-0.778$ & 25944.0 & $-0.533$ \\
GPT-5           & 16796.0 & $-0.816$ & 36438.5 & $-0.769$ & 24748.0 & $-0.462$ \\
GPT-OSS 120B    & 16388.0 & $-0.772$ & 33292.0 & $-0.616$ & 23736.0 & $-0.402$ \\
Gemini 2.5 Pro  & 14484.0 & $-0.566$ & 34205.5 & $-0.660$ & 1052.0  & $-0.429^\dagger$ \\
Gemini 2.5 Flash & 15368.0 & $-0.662$ & 34205.5 & $-0.660$ & 24288.0 & $-0.435$ \\
Kimi K2         & 15844.0 & $-0.713$ & 33799.5 & $-0.640$ & 11672.0 & $-0.442$ \\
Qwen3 235B      & 15776.0 & $-0.706$ & 33089.0 & $-0.606$ & 23644.0 & $-0.397$ \\
DeepSeek V3     & 16932.0 & $-0.831$ & 36235.5 & $-0.759$ & 25668.0 & $-0.516$ \\
Mistral 3.2 24B & 16660.0 & $-0.802$ & 36641.5 & $-0.778$ & 26772.0 & $-0.582$ \\
Mistral Nemo    & 15776.0 & $-0.706$ & 31262.0 & $-0.517$ & 25576.0 & $-0.511$ \\
Llama 4 Maverick & 16660.0 & $-0.802$ & 36844.5 & $-0.788$ & 26864.0 & $-0.587$ \\
Llama 3.3 70B   & 8738.0  & $-0.691$ & 34104.0 & $-0.655$ & 25024.0 & $-0.478$ \\
Llama 3.1 8B    & 15572.0 & $-0.684$ & 30856.0 & $-0.498$ & 24380.0 & $-0.440$ \\
\bottomrule
\end{tabular}}
\caption{Mann-Whitney U statistics and rank-biserial effect sizes ($r$) for the Easy vs.\ Hard distractor comparison across all 13 models and 3 domains. All $p < 0.0001$ except $^\dagger$Gemini 2.5 Pro on PFG  ($p = 0.012$). Negative $r$ values indicate higher scores on the Easy Set; $|r| \geq 0.3$ is considered a large effect.}
\label{tab:mannwhitney_full}
\end{table*}

\section{Prompt Templates}
\label{appendix:prompts}

\subsection{Preliminary Prompts}
This section includes prompts that are either used in parts, or to derive summaries that are later used in the actual task prompts.

\begin{promptbox}{Mental State Questions}
The mental state questions in the proceeding prompts are taken from the following questions: \\

Belief: What do you believe in? \\
Desires: What do you want to do? \\
Intentions: What will you do? \\
Emotions: How do you feel? \\
Knowledge: What do you know? \\
Trust: Do you trust the \textcolor{blue}{\{initiator role\}}?
\end{promptbox}

\begin{promptbox}{Mental State Prefix Format}
The mental state answers in the proceeding prompts follow the given format: \\

Belief: I believe \\
Desires: I want \\
Intentions: I will \\
Emotions: I feel \\
Knowledge: I know \\
Trust: I view the \textcolor{blue}{\{initiator role\}} as
\end{promptbox}

\begin{promptbox}{Summary of conversation}
Generate a summary of the conversation context in less than 50 words. \\

Conversation Context: \\
\textcolor{blue}{\{conversation\}}
\end{promptbox}

\subsection{Data Generation}

\begin{promptbox}{Optimal Mental State and Distractors}
Assume you are a \textcolor{blue}{\{recipient role\}} discussing with a skilled \textcolor{blue}{\{initiator role\}} \textcolor{blue}{\{topic type\}}. Your current task is to think about your own mental states at the current state based on the provided context and conversation. The current state of the \textcolor{blue}{\{session type\}} session is as follows: \\

\textcolor{blue}{\{conversation\}} \\

You need to infer your own mental states from your perspective as the \textcolor{blue}{\{recipient role\}}. Answer the following questions for the \textcolor{blue}{\{mental state\}} state. Generate 4 different answers for the mental state provided below in the form of questions. Your first answer will always represent your most accurate answer. The other answers should be wrong but confusion inducing. \\

\textcolor{blue}{\{mental state\}}: \textcolor{blue}{\{mental state question\}} \\

Use the following prefix format for \textcolor{blue}{\{mental state\}}: \\
\textcolor{blue}{\{mental state\}}: \textcolor{blue}{\{mental state prefix format\}}

\end{promptbox}

\begin{promptbox}{Dialogue Trajectory Distractors}
Given the following conversation between two agents: \\

\textcolor{blue}{\{conversation\}} \\

Here, the \textcolor{blue}{\{recipient role\}} has the following mental state: \textcolor{blue}{\{correct mental state\}} \\

The two future utterances after this conversation based on the above mental state are: \\
\textcolor{blue}{\{correct dialogue trajectory\}} \\

Generate an alternative conversation direction, each direction should follow the same format as the utterances provided just before, for the following mental state: \\
\textcolor{blue}{\{distractor mental states\}} \\

This conversation direction should be objectively different from the original utterances provided. strictly follow the format given below for generating the alternative conversation direction. Do not generate any other text or explanation, just the alternative conversation direction. \\

Format: \\
\textcolor{blue}{\{recipient role\}}: \textcolor{blue}{\{dialogue 1\}} \\
\textcolor{blue}{\{initiator role\}}: \textcolor{blue}{\{dialogue 2\}} \\
\textcolor{blue}{\{recipient role\}}: \textcolor{blue}{\{dialogue 3\}} \\
\textcolor{blue}{\{initiator role\}}: \textcolor{blue}{\{dialogue 4\}}
\end{promptbox}

\begin{promptbox}{Counterfactuals}
You are an expert in Theory of Mind reasoning and generating counterfactual statements. \\

Task: \\
You will be provided with an excerpt from a conversation along with the current \textcolor{blue}{\{state\}} state of \textcolor{blue}{\{recipient role\}}. \\

You need to generate a counterfactual mental state that is obviously not true at all given the current conversation. It should be CONTRARY to the provided state. \\

Conversation context: \\
\textcolor{blue}{\{conversation\}} \\

Mental State: \\
\textcolor{blue}{\{current mental state\}} \\

Instruction: \\
Output only the counterfactual. Do not add explanations or other verbosity.
Your output should strictly follow the structure "\textcolor{blue}{\{mental state prefix format\}}". NO FORMATTING NEEDS TO BE DONE. \\

Counterfactual:
\end{promptbox}

\subsection{Main Benchmarking Tasks}

\begin{promptbox}{Retrospective}
You are an expert in Theory of Mind reasoning. \\

Task: \\
You will be provided with conversation between two agents \textcolor{blue}{\{initiator role\}} and \textcolor{blue}{\{recipient role\}} engaging in a \textcolor{blue}{\{session type\}} session on the topic of \textcolor{blue}{\{topic type\}}. \\

Your goal is to correctly infer \textcolor{blue}{\{recipient role\}}'s mental state, based on the above conversation. You will be provided with a set of options, and you need to choose the most appropriate one that reflects the mental state. \\
The correct option must be consistent with the provided conversation context. \\

Conversation Context: \\
\textcolor{blue}{\{conversation\}} \\

Mental State Options: \\
\textcolor{blue}{\{options\}} \\

Instruction: \\
Output only the letter of the correct option (e.g., "A", "B", "C", or "D"). Do not add explanations or other verbosity. \\
Your output should be strictly one of: A, B, C, D. NO FORMATTING NEEDS TO BE DONE.  \\

Answer:
\end{promptbox}

\begin{promptbox}{Prospective}
You are an expert in Theory of Mind reasoning. \\

Task: \\
You will be provided with internal Mental State profile (Belief, Desire, Intention, Emotion, Knowledge, Trust) of \textcolor{blue}{\{recipient role\}} during the conversation. \\

Your goal is to identify which of the candidate conversation segments is the most plausible continuation of this conversation. \\
The correct option must be consistent with the provided Mental States of the \textcolor{blue}{\{recipient role\}}. \\

Mental state of \textcolor{blue}{\{recipient role\}} \\
\textcolor{blue}{\{mental states\}} \\

Candidate Conversation Segments \\
\textcolor{blue}{\{options\}} \\

Instruction: \\
Output only the letter of the correct option (e.g., "A", "B", "C", or "D"). Do not add explanations or other verbosity. \\
Your output should be strictly one of: A, B, C, D. NO FORMATTING NEEDS TO BE DONE. \\

Answer:
\end{promptbox}

\begin{promptbox}{LLM-as-a-Judge Evaluation}
You are an expert linguistic evaluator. Your task is to perform a dialogue-anchored semantic evaluation, comparing an AI-generated mental state inference against a Human-written gold standard. \\

IMPORTANT RULES \\
1. The Human Inference is the absolute gold standard. Do not evaluate whether the Human Inference is correct or accurate. \\ 
2. Your only job is to evaluate how closely the AI Inference aligns with the Human's proposition, causal reasoning, and specificity. \\ 
3. Use the Source Dialogue strictly as a reference to understand what the Human was responding to, so you can check if the AI responded to the exact same conversational triggers. \\
4. Ignore word overlap, grammar, and phrasing. Evaluate purely on conceptual and semantic alignment. \\

INPUTS \\
Source Dialogue: \\
\textcolor{blue}{\{conversation\}} \\

Human Inference: \\
\textcolor{blue}{\{human written inference\}} \\

AI Inference: \\
\textcolor{blue}{\{llm generated inference\}} \\

EVALUATION RUBRICS \\

Evaluate the AI Inference on a scale of 0 to 3 for the following three dimensions: \\

1. Propositional Agreement (0-3) \\
Do the AI and human inferences express the same core proposition about the mental state? \\
- 3: Same core claim — they agree on what the speaker believes/desires/intends, even if worded differently. \\
- 2: Partial agreement — they agree on the general direction, but the AI's inference is either more narrow or more broad than the human's. \\
- 1: Tangential — they are about the same topic but express meaningfully different claims. \\
- 0: Contradiction or complete mismatch. \\

2. Causal Attribution Agreement (0-3) \\
Do the AI and human anchor the mental state to the same conversational cause or evidence? \\
- 3: Perfect match — the AI identifies the same triggering utterance, event, or pattern in the dialogue as the human's inference implies. \\
- 2: Related match — the AI identifies a related but not identical cause (plausible, but not exactly what the human is responding to). \\
- 1: Weak match — the AI's causal anchor is vague or only weakly connected to what the human appears to be responding to. \\
- 0: Mismatch — the AI attributes the mental state to something the human's inference does not recognize as relevant. \\

3. Specificity Alignment (0-3) \\
Are the AI and human inferences operating at the same level of specificity/granularity? \\
- 3: Perfect Granularity Match — both operate at the exact same level of detail. \\
- 2: Minor Detail Mismatch — they share the same core scope, but one includes a minor contextual detail that the other omits. \\
- 1: Severe Granularity Mismatch — one makes a highly individuated, context-specific claim, while the other relies on a sweeping generalization. \\
- 0: Complete Divergence in Scope — the inferences exist on entirely different planes. \\

OUTPUT FORMAT \\
You must output a valid JSON object. Provide a brief 1-sentence rationale for each score before providing the integer. 
\end{promptbox}

\begin{promptbox}{Reasoning Trace Injection Pilot}
You are an expert in Theory of Mind reasoning. \\

Task: \\
You will be provided with internal Mental State profile (Belief, Desire, Intention, Emotion, Knowledge, Trust) of \textcolor{blue}{\{recipient role\}} during the conversation. \\

Your goal is to identify which of the candidate conversation segments is the most plausible continuation of this conversation. \\
The correct option must be consistent with the provided Mental States of the \textcolor{blue}{\{recipient role\}}. \\

Mental state of \textcolor{blue}{\{recipient role\}} \\
\textcolor{blue}{\{all mental states\}} \\

Candidate Conversation Segments \\
\textcolor{blue}{\{trajectory options\}} \\

You may use the following reasoning trace from an expert to guide your answer: \\
\textcolor{blue}{\{reasoning trace\}} \\

Instruction \\
Output only the letter of the correct option (e.g., "A", "B", "C", or "D"). Do not add explanations or other verbosity. \\
Your output should be strictly one of: A, B, C, D. NO FORMATTING NEEDS TO BE DONE. \\

Answer:
\end{promptbox}

\subsection{Dialogue Context Degradation Analysis}

\begin{promptbox}{State only}
You are an expert in Theory of Mind reasoning. \\

Task: \\
You will be provided with internal Mental State profile (Belief, Desire, Intention, Emotion, Knowledge, Trust) of \textcolor{blue}{\{recipient role\}}  during the conversation. \\

Your goal is to identify which of the candidate conversation segments is the most plausible continuation of this conversation. \\
The correct option must be consistent with the provided Mental States of the \textcolor{blue}{\{recipient role\}}. \\

Mental State of \textcolor{blue}{\{recipient role\}} \\
\textcolor{blue}{\{all mental states\}} \\

Candidate Conversation Segments \\
\textcolor{blue}{\{trajectory options\}} \\

Instruction \\
Output only the letter of the correct option (e.g., "A", "B", "C", or "D"). Do not add explanations or other verbosity. \\
Your output should be strictly one of: A, B, C, D. NO FORMATTING NEEDS TO BE DONE. \\

Answer:
\end{promptbox}

\begin{promptbox}{Summary + States}
You are an expert in Theory of Mind reasoning. \\

Task: \\
You will be provided with internal Mental State profile (Belief, Desire, Intention, Emotion, Knowledge, Trust) of \textcolor{blue}{\{recipient role\}}  during the conversation. \\

Your goal is to identify which of the candidate conversation segments is the most plausible continuation of this conversation. \\
The correct option must be consistent with the provided Mental States of the \textcolor{blue}{\{recipient role\}}. \\

Conversation Context Summary: \\
\textcolor{blue}{\{conversation summary\}} \\

Mental State of \textcolor{blue}{\{recipient role\}} \\
\textcolor{blue}{\{all mental states\}} \\

Candidate Conversation Segments \\
\textcolor{blue}{\{trajectory options\}} \\

Instruction \\
Output only the letter of the correct option (e.g., "A", "B", "C", or "D"). Do not add explanations or other verbosity. \\
Your output should be strictly one of: A, B, C, D. NO FORMATTING NEEDS TO BE DONE. \\

Answer:
\end{promptbox}

\begin{promptbox}{Conversation History + States}
You are an expert in Theory of Mind reasoning. \\

Task: \\
You will be provided with internal Mental State profile (Belief, Desire, Intention, Emotion, Knowledge, Trust) of \textcolor{blue}{\{recipient role\}}  during the conversation. \\

Your goal is to identify which of the candidate conversation segments is the most plausible continuation of this conversation. \\
The correct option must be consistent with the provided Mental States of the \textcolor{blue}{\{recipient role\}}. \\

Conversation Context: \\
\textcolor{blue}{\{conversation\}} \\

Mental State of \textcolor{blue}{\{recipient role\}} \\
\textcolor{blue}{\{all mental states\}} \\

Candidate Conversation Segments \\
\textcolor{blue}{\{trajectory options\}} \\

Instruction \\
Output only the letter of the correct option (e.g., "A", "B", "C", or "D"). Do not add explanations or other verbosity. \\
Your output should be strictly one of: A, B, C, D. NO FORMATTING NEEDS TO BE DONE. \\

Answer:
\end{promptbox}







\subsection{Qualitative Pairwise Traces Analysis}

\begin{promptbox}{LLM-as-a-Judge Evaluation}
You are an expert AI evaluator assessing reasoning traces produced by language models in Theory of Mind and Mental State reasoning tasks.\\

Your task is to evaluate two reasoning traces (Trace\_A and Trace\_B) independently against a 4-dimension rubric. You will be provided with:\\
1. Domain\\
2. Mental State Profile (containing up to six BDIEKT attributes: Belief, Desire, Intention, Emotion, Knowledge, Trust)\\
3. Correct Option\\
4. Distractor Options (Distractor 1, Distractor 2, Distractor 3)\\
5. Trace\_A and Trace\_B (masked reasoning traces; you are not told which model generated which trace)\\

Score each trace strictly based on its content according to the following rubric (0-2 for each dimension, accompanied by a single-sentence justification per score):\\
\\
--- RUBRIC ---\\
\\
1. Attribute Coverage\\
- 0 = references only 1 profile attribute or none explicitly\\
- 1 = references 2-3 attributes\\
- 2 = explicitly references 4+ of the six BDIEKT attributes (Belief, Desire, Intention, Emotion, Knowledge, Trust) in its reasoning\\
\\
2. Elimination Specificity\\
- 0 = rejects options without stating why, or with vague language like "doesn't fit as well"\\
- 1 = some specific reasons given, inconsistently\\
- 2 = each rejected option is tied to a specific attribute-level conflict\\
\\
3. Consistency Verification\\
- 0 = stops reasoning once a plausible option is found\\
- 1 = partial cross-checking against some attributes\\
- 2 = explicitly verifies the selected option against most/all stated attributes before concluding
(Note: Use knowledge of the Correct\_Option ONLY to assess whether the trace's own verification process was sound. Do not score a trace higher simply because it happened to reach the correct answer.)\\
\\
4. Surface vs. Structural Language\\
- 0 = reasoning relies primarily on topical/thematic similarity language\\
- 1 = a mix of topical and structural language\\
- 2 = reasoning is framed primarily in terms of logical/structural consistency with specific stated attributes\\
\\
Do not consider whether the final answer was correct --- score ONLY the reasoning process itself.\\
\\
After scoring both traces, write one sentence describing the single most notable difference between Trace\_A's and Trace\_B's reasoning approach, if any is apparent.\\
\\
--- EVALUATION ITEM ---\\
- Domain: \textcolor{blue}{\{domain\}}\\
- Mental State Profile: \textcolor{blue}{\{mental\_state\_profile\}}\\
- Correct Option: \textcolor{blue}{\{correct\_option\}}\\
- Distractor 1:\textcolor{blue}{\{distractor\_1\}}\\
- Distractor 2: \textcolor{blue}{\{distractor\_2\}}\\
- Distractor 3: \textcolor{blue}{\{distractor\_3\}}\\
\\
--- TRACE A ---\\
\textcolor{blue}{\{trace\_a\}}\\
\\
--- TRACE B ---\\
\textcolor{blue}{\{trace\_b\}}\\
\\
Please perform the qualitative evaluation of Trace A and Trace B based strictly on the rubric and output the JSON response.

\end{promptbox}

\section{Task Examples}
\label{appendix:task_examples}
To illustrate the structure and complexity of the DialToM benchmark, we provide representative examples of the Retrospective Inference and Prospective Diagnostic Forecasting tasks in this section. Each example is labeled with a unique Task ID, corresponding to its specific instance in the DialToM dataset.

\subsection{Retrospective Inference}
We include three examples of Retrospective Inference tasks (Literal ToM), one for each conversation domain: Counseling, Emotional Support, and Persuasion. These examples demonstrate the model's ability to map multi-turn conversation contexts to specific mental state attributes

\begin{promptbox}{Retrospective Task 1: Counseling (ID: 17)}
You are an expert in Theory of Mind reasoning. \\

Task: \\
You will be provided with conversation between two agents supporter and seeker engaging in a Counseling Session on the topic of on reducing alcohol consumption. \\

Your goal is to correctly infer seeker's \textbf{Emotional} state, based on the conversation below. You will be provided with a set of options, and you need to choose the most appropriate one that reflects the Emotion state. \\

The correct option must be consistent with the provided conversation context. \\

Conversation Context: \\
    \textbf{counselor}: And with friends, you feel like it's easier to-to hang out with alcohol, or maybe you wouldn't have as much fun, or your friends wouldn't like you as much or?\\
    \textbf{client}: I guess, I mean, since it's just there.\\
    \textbf{counselor}: Okay.\\
    \textbf{client}: So-\\
    \textbf{counselor}: Got it.\\
    \textbf{client}: -it ties in with the whole belonging thing.\\
    \textbf{counselor}: Okay. Well, when you think about your drinking and, sort of, looking forward in 5 years or 10 years, how do you think that drinking might fit in with your life?\\
    \textbf{client}: Um, it probably will have a very small role, if any-\\
    \textbf{counselor}: Mm-hmm.\\
    \textbf{client}: -in my life, considering I have big plans for my future. So.\\
    \textbf{counselor}: Okay. Can you expand on that a little bit more? What do you mean?\\
    \textbf{client}: I mean, I don't think I would be, you know, drinking as much or going to parties as much if I'm studying for school, or in college, or trying to get in law school and things like that.\\
    \textbf{counselor}: Okay, so, the-the alcohol is something that might make it harder to reach your goals?\\
    \textbf{client}: Yeah, definitely.\\

Mental State Options:\\
A: I feel indifferent about my drinking habits right now.\\
B: I feel excited about drinking because it’s a big part of my social life.\\
C: I feel anxious about the idea of not drinking with friends.\\
D: I feel determined to reduce my alcohol consumption because I have clear long-term goals.\\

Instruction:\\
Output only the letter of the correct option (e.g., "A", "B", "C", or "D"). Do not add explanations or other verbosity.\\
Your output should be strictly one of: A, B, C, D. NO FORMATTING NEEDS TO BE DONE. ONLY OUTPUT THE OPTION AND NOTHING ELSE. YOUR OUTPUT SHOULD STRICTLY BE ONE OF A, B, C, or D.\\

Answer:
\end{promptbox}

\begin{promptbox}{Retrospective Task 2: Emotional Support (ID: 22)}
You are an expert in Theory of Mind reasoning. \\

Task: \\
You will be provided with conversation between two agents supporter and seeker engaging in an Emotional Support Conversation session on the topic of on anger over breakup with partner. \\

Your goal is to correctly infer seeker's \textbf{Knowledge} state, based on the conversation below. You will be provided with a set of options, and you need to choose the most appropriate one that reflects the Knowledge state. \\

The correct option must be consistent with the provided conversation context. \\

Conversation Context: \\
    \textbf{seeker}: It has been happening for at least 6 months now and it usually happens every weekend \\
    \textbf{supporter:} I can sense your frustration. It must be overwhelming at times and make you dread the weekends. It sounds like it has become a habit for your ex. It might help if you write a letter to them explaining that you are ready to move on and are asking them to do the same. Explain that you are giving them the opportunity to stop harassing you in the following ways [list some of the details you have documented]. Inform them that if they do not stop, you are considering taking it further regarding harassing communications and that this letter serves as your intention to no longer tolerate their actions. Send it via certified mail and keep a copy. This will give them no way to immediately respond and would be the first step of official documentation should they decide to continue.  Do you think that might help? \\
    \textbf{seeker:} I think that is an excellent idea! That is a good way to end this once and for all.\\
    \textbf{supporter:} I sincerely hope it is a step toward you regaining your control  over how you are able to interact with people. I wish you the best in dealing with this and hope you begin to feel better about it soon, especially entering into a new year!\\
    \textbf{seeker:} Thank you very much!\\

Mental State Options:\\
A: I know that responding impulsively in anger can resolve the situation more gently.\\
B: I know that avoiding any communication is the best way to make the situation go away completely.\\
C: I know that sending a well-documented letter can serve as both communication and legal evidence to stop harassment.\\
D: I know that ignoring the problem will make it disappear without any further action required.\\

Instruction:\\
Output only the letter of the correct option (e.g., "A", "B", "C", or "D"). Do not add explanations or other verbosity.\\
Your output should be strictly one of: A, B, C, D. NO FORMATTING NEEDS TO BE DONE. ONLY OUTPUT THE OPTION AND NOTHING ELSE. YOUR OUTPUT SHOULD STRICTLY BE ONE OF A, B, C, or D.\\

Answer:
\end{promptbox}

\begin{promptbox}{Retrospective Task 3: Persuasion (ID: 19)}
You are an expert in Theory of Mind reasoning. \\

Task: \\
You will be provided with conversation between two agents supporter and seeker engaging in a Persuasion Conversation session on Persuading the persuadee to donate to SaveTheChildren organization.\\

Your goal is to correctly infer persuadee's \textbf{Intention} state, based on the conversation below. You will be provided with a set of options, and you need to choose the most appropriate one that reflects the Intention state. \\

The correct option must be consistent with the provided conversation context. \\

Conversation Context: \\
    \textbf{persuader}: Hello!\\
    \textbf{persuadee}: Hi there!\\
    \textbf{persuader}: how are things going for you?\\
    \textbf{persuadee}: They are going well - are we supposed to decide about a donation out of our \$2?\\
    \textbf{persuader}: Yes we are. how do you feel about it?\\
    \textbf{persuadee}: Well it would depend on what the donation is supposed to go to.\\
    \textbf{persuader}: The charity is Save The Children.  It promotes  children's rights, provides relief and helps support children in developing countries.\\
    \textbf{persuadee}: Oh OK well that sounds like a good cause for a donation.\\
    \textbf{persuader}: Excellent! I thought so as well!\\
    \textbf{persuadee}: I was thinking that donating half of the \$2 would be a fair amount.\\
    \textbf{persuader}: That's what I was thinking as well. If we donate half, we still got something in return.\\
    \textbf{persuadee}: I think that sounds like a good plan and it is fair to everyone.\\
    \textbf{persuader}: Excellent. I will donate half and do will do the same then?  that's 1 dollar each for donation.\\
    \textbf{persuadee}: Yes! I have been hearing of this charity since I was very young.\\

Mental State Options:\\
A: I will donate the full \$2 to Save The Children.\\
B: I will donate \$1 to Save The Children.\\
C: I will wait and not make a decision yet.\\
D: I will save the entire \$2 for myself.\\

Instruction:\\
Output only the letter of the correct option (e.g., "A", "B", "C", or "D"). Do not add explanations or other verbosity.\\
Your output should be strictly one of: A, B, C, D. NO FORMATTING NEEDS TO BE DONE. ONLY OUTPUT THE OPTION AND NOTHING ELSE. YOUR OUTPUT SHOULD STRICTLY BE ONE OF A, B, C, or D.\\

Answer:
\end{promptbox}

\subsection{Prospective Diagnostic Forecasting}
We present three examples of Prospective Diagnostic Forecasting (Functional ToM) across the same domains, highlighting the state-driven nature of our diagnostic probing.

\begin{promptbox}{Prospective Task 1: Counseling (ID: 17)}
You are an expert in Theory of Mind reasoning. \\

Task: \\
You will be provided with internal Mental State profile (Belief, Desire, Intention, Emotion, Knowledge, Trust) of client during the conversation. \\

Your goal is to identify which of the candidate conversation segments is the most plausible continuation of this conversation. \\
The correct option must be consistent with the provided Mental States of the client. \\

Mental state of client \\
Belief: I believe that drinking is risky but it's hard to avoid because of social situations.\\
Desires: I want to fit in with my friends when we're at parties.\\
Intentions: I will consider how my drinking habits might affect my relationships with my family and friends.\\
Emotions: I feel conflicted because I enjoy socializing with my friends but don't want to disappoint my family.\\
Knowledge: I know that underage drinking is risky and illegal.\\
Trust: I view the counselor as someone who genuinely cares and seems understanding, so I trust them to talk about my drinking habits.\\

Candidate Conversation Segments\\
A:\\
\textbf{counselor}: And with friends, you feel like it's easier to-to hang out with alcohol, or maybe you wouldn't have as much fun, or your friends wouldn't like you as much or?\\
\textbf{client}: I guess, I mean, since it's just there.\\
\textbf{counselor}: Okay.\\
\textbf{client}: So-\\

B:\\ 
\textbf{counselor}: And do you ever feel like saying no to drinking is hard because of peer pressure?\\
\textbf{client}: Sometimes, yeah.\\
\textbf{counselor}: What do you think would help you feel more comfortable to say no?\\
\textbf{client}: I guess if more friends felt the same way.\\

C:\\ 
\textbf{counselor}: How do you feel when you're in situations without alcohol?\\
\textbf{client}: I guess sometimes it's awkward, but it's fine.\\
\textbf{counselor}: Okay, so it's not something you necessarily need to enjoy yourself?\\
\textbf{client}: Yeah, I can still have fun without it.

D:\\
\textbf{counselor}: What are some things you enjoy doing with your friends that don't involve alcohol?\\
\textbf{client}: We like just hanging out, watching movies or playing games.\\
\textbf{counselor}: That sounds fun. Do you feel you can still connect with them without the drinking aspect?\\
\textbf{client}: Yeah, I think so. It's more about being together, really.\\

Instruction:\\
Output only the letter of the correct option (e.g., "A", "B", "C", or "D"). Do not add explanations or other verbosity. \\
Your output should be strictly one of: A, B, C, D. NO FORMATTING NEEDS TO BE DONE.\\

Answer:
\end{promptbox}

\begin{promptbox}{Prospective Task 2: Emotional Support (ID: 22)}
You are an expert in Theory of Mind reasoning. \\

Task: \\
You will be provided with internal Mental State profile (Belief, Desire, Intention, Emotion, Knowledge, Trust) of client during the conversation. \\

Your goal is to identify which of the candidate conversation segments is the most plausible continuation of this conversation. \\
The correct option must be consistent with the provided Mental States of the client. \\

Mental state of client \\
Belief: I believe my ex is struggling to accept the breakup and it's affecting my mental peace.\\
Desires: I want to find a way to stop my ex from continually reaching out and creating new accounts.\\
Intentions: I will consider blocking them and encourage my friends to do the same to create some distance and peace.\\
Emotions: I feel overwhelmed and exhausted by the situation with my ex.\\
Knowledge: I know that my ex's actions are overwhelming and they are struggling to let go.\\
Trust: I view the supporter as understanding and helpful, so yes, I trust them.\\

Candidate Conversation Segments\\
A:\\
\textbf{seeker}: It has been happening for at least 6 months now and it usually happens every weekend\\
\textbf{supporter}: I can sense your frustration. It must be overwhelming at times and make you dread the weekends. It sounds like it has become a habit for your ex. It might help if you write a letter to them explaining that you are ready to move on and are asking them to do the same. Explain that you are giving them the opportunity to stop harassing you in the following ways [list some of the details you have documented]. Inform them that if they do not stop, you are considering taking it further regarding harassing communications and that this letter serves as your intention to no longer tolerate their actions. Send it via certified mail and keep a copy. This will give them no way to immediately respond and would be the first step of official documentation should they decide to continue.  Do you think that might help?\\
\textbf{seeker}: I think that is an excellent idea! That is a good way to end this once and for all.\\
\textbf{supporter}: I sincerely hope it is a step toward you regaining your control  over how you are able to interact with people. I wish you the best in dealing with this and hope you begin to feel better about it soon, especially entering into a new year!\\

B:\\ 
\textbf{seeker}: It's been ongoing for around 6 months, and I notice it happens mostly when I've been out with friends.\\
\textbf{supporter}: That must be hard to always feel like you're being watched. Have you considered documenting each incident, including witness accounts, for your own peace of mind and to keep a clear timeline of events? It could be helpful if you decide to take further steps later on. Is that something you think could be useful?\\
\textbf{seeker}: That's a good idea. I'll start keeping better track of everything.\\
\textbf{supporter}: It sounds like a solid plan. After documenting, you might also feel more in control and less anxious. If you need any help organizing it, let me know. Your well-being is important, and having support in this can make a big difference.\\

C:\\ 
\textbf{seeker}: They pace up and down our street weekly and it really makes me anxious.\\
\textbf{supporter}: It's understandable to feel on edge. It might be beneficial to consider speaking with a professional for guidance. They can offer strategies to emotionally distance yourself from your ex's actions.\\
\textbf{seeker}: I've thought about reaching out for professional help, but I'm hesitant.\\
\textbf{supporter}: It's normal to feel hesitant. Sometimes an initial consultation can provide clarity on whether it's the right path for you. You deserve peace of mind.\\

D:\\
\textbf{seeker}: It has been happening for about half a year, and it tends to intensify over the weekends.\\
\textbf{supporter}: That sounds really tough. Weekends should be a time for you to relax, not worry. Have you considered talking to a professional about setting clearer boundaries or reaching out to support groups who may have dealt with similar situations? Sometimes sharing experiences with others who understand can offer additional support and strategies.\\
\textbf{seeker}: That sounds like a good suggestion. It might be helpful to talk to someone who has been through something similar.\\
\textbf{supporter}: Definitely. Listening to others who have navigated this path can provide insights and make you feel less isolated. I hope you find some useful perspectives and that it helps bring some peace back to your weekends.\\

Instruction:\\
Output only the letter of the correct option (e.g., "A", "B", "C", or "D"). Do not add explanations or other verbosity. \\
Your output should be strictly one of: A, B, C, D. NO FORMATTING NEEDS TO BE DONE.\\

Answer:
\end{promptbox}

\begin{promptbox}{Prospective Task 3: Persuasion (ID: 19)}
You are an expert in Theory of Mind reasoning. \\

Task: \\
You will be provided with internal Mental State profile (Belief, Desire, Intention, Emotion, Knowledge, Trust) of client during the conversation. \\

Your goal is to identify which of the candidate conversation segments is the most plausible continuation of this conversation. \\
The correct option must be consistent with the provided Mental States of the client. \\

Mental state of client \\
Belief: I believe that donating to Save The Children is a meaningful and impactful way to support children's rights and welfare globally.\\
Desires: I want to donate half of the \$2 to Save The Children because it supports a good cause.\\
Intentions: I will donate \$1 to Save The Children.\\
Emotions: I feel positive about donating to Save the Children because it's a cause I believe in.\\
Knowledge: I know that Save The Children is a charity focused on supporting children's rights and providing relief in developing countries.\\
Trust: I view the persuader as trustworthy because they clearly explained the charity's mission and showed enthusiasm for the cause.\\

Candidate Conversation Segments\\
A:\\
\textbf{persuader}: I have donated there in the past, and I know the good work that they do here and abroad.\\
\textbf{persuadee}: Our children's group in my childhood church used to take up coins to donate to them.\\
\textbf{persuader}: That's really awesome!  With what they accomplish with a little money, I feel we will make a small difference.\\
\textbf{persuadee}: I hope so - I've seen the ads on television and some of those poor children have to live in terrible conditions.\\

B:\\
\textbf{persuader}: I'm really glad we’re putting our money toward an organization that truly makes an impact.\\
\textbf{persuadee}: I agree, it feels reassuring knowing our contribution is going somewhere meaningful.\\
\textbf{persuader}: Even small amounts can add up and create real change for people who need it.\\
\textbf{persuadee}: Definitely, and it’s inspiring to think that our choice might encourage others to give too.\\

C:\\
\textbf{persuader}: I actually volunteered at one of their fundraising events last year, and it was a really meaningful experience.\\
\textbf{persuadee}: Oh wow, that's amazing! Have you been involved in any of their other charity events too?\\
\textbf{persuader}: Not beyond that one, but seeing the impact up close was incredible. It really showed me how much good they’re able to do.\\
\textbf{persuadee}: That’s so inspiring. I’ve read stories about how they help rebuild schools, and it’s encouraging to hear that the impact is real firsthand.\\

D:\\
\textbf{persuader}: It really feels powerful knowing that even a small amount from us could help create real change.\\
\textbf{persuadee}: I agree. I've come across stories about how their programs have completely uplifted whole communities.\\
\textbf{persuader}: That’s so encouraging to hear. Maybe down the line we could even inspire more people to contribute too.\\
\textbf{persuadee}: I love that idea. If more of us get involved, the impact could grow and reach even more children who need it.\\

Instruction:\\
Output only the letter of the correct option (e.g., "A", "B", "C", or "D"). Do not add explanations or other verbosity. \\
Your output should be strictly one of: A, B, C, D. NO FORMATTING NEEDS TO BE DONE.\\

Answer:
\end{promptbox}

\subsection{Gemini 3 Pro's Reasoning Traces}
\label{appendix:reasoning_traces_examples}
We present representative examples of Gemini 3 Pro's truncated internal reasoning traces from the MI domain, which were injected into student models in Section~\ref{subsec:reasoning_injection}. As shown in the traces below, while the final five sentences containing terminal answer declarations and summary conclusions were strictly truncated to prevent direct label cheating, the generated pathways naturally rely on a systematic, verbal process-of-elimination rationale (e.g., explicitly naming, comparing, and dismissing option identifiers like ``Option A'' or ``Option B'' during the analytical progression). 

These excerpts transparently demonstrate that Gemini 3 Pro solves the Prospective task by running an active, comparative state-to-text matching matrix across all available candidates. Consequently, the performance gains reported in Section~\ref{subsec:reasoning_injection} reflect the student models' ability to ingest and execute this long-context process-of-elimination logic.

\begin{promptbox}{MI Reasoning Trace 1; C is Correct}
**Analysis of Conversation Segment Plausibility** \\

Okay, so I'm presented with a client's mental state and need to identify the most fitting conversation snippet. I have to leverage my understanding of the client's beliefs, desires, intentions, emotions, and knowledge – and also their trust in the counselor. My initial thought process revolves around matching the nuances of the client's internal world to the available dialogue options. \\

First, I zero in on the core of the client's belief: cutting back on fruit juice is linked to weight loss. They *desire* a balance, wanting to enjoy OJ but also shed weight. They *intend* to reduce intake gradually and look for healthier alternatives. The *emotions* are key too – surprise and a realization about the juice's impact. Finally, there's the *knowledge* that excessive juice leads to weight gain, and the crucial element of *trust* in the counselor's ability to provide personalized advice. \\

Now, let's analyze each conversation candidate. [MASK]. This hits on the *desire* for balance, as the client wants to enjoy OJ while finding a weight-loss solution. [MASK]. It aligns with the *intention* to find alternatives, but the *desire* is more about balancing current juice consumption. [MASK]. This approach uses the client's knowledge that they ""need"" to lose weight, and therefore matches their *belief* that cutting back *will* help. [MASK]. \\

The prompt highlights a moment of realization by the client. The counselor's role now is to leverage this newfound awareness. One option [MASK] stands out. It's the most effective in eliciting “Change Talk.” The specific phrasing – [MASK] – directly reflects the core belief that cutting back is linked to weight loss. The client’s answer in [MASK] is [MASK], which directly echoes the ""Belief"" element within the client's provided mental state. \\

Thinking about the ""Desires,"" [MASK] is a solid fit, particularly for those desiring to balance consumption with weight loss. However, usually, in datasets such as those used in motivational interviewing, the mental state description represents the *interpretation* of the correct conversation segment. The ""Belief"" in the mental state description almost directly mirrors the client's response in [MASK]. Also, the phrasing ""personalized advice"" implies the counselor is validating the client's motivation, which is present in [MASK]. \\
\end{promptbox}

\begin{promptbox}{MI Reasoning Trace 2; D is Correct}
**Choosing the Right Conversation Path** \\

Okay, here's what I'm thinking. The core of this is about connecting with the client's current state. I need to figure out which conversation segment best resonates with where they are right now. The client's mental state is pretty clear, and it’s all pointing towards asthma management, even though general overwhelm is also present. \\

Looking at the options, it's pretty easy to discount [MASK]. They all immediately steer the conversation toward [MASK], which just isn't what's on the table, based on what the client is explicitly trying to work on. They are trying to organize their thoughts around asthma management, as I see it. It's the whole point. The client *intends* to make a list specifically relating to their asthma, which is where their Knowledge and Trust are focused, and that is their core Desire. So, when the conversation veers off, as it does in [MASK], it’s not clicking with what they are trying to do. \\

But [MASK] – that’s different. [MASK] allows a sense of security to emerge. They agree, expressing the simple goal of organizing things, ""putting pen to paper."" The counselor's initial acceptance of the goal provides the space needed. Then, *finally*, the client verbalizes, ""Um, okay, so my asthma."" That's the key. That’s the explicit connection to their *Intention*, their *Desire*, and the *Knowledge* they are bringing to the table, and the ""Um, okay"" matches their Emotion of being unsure where to begin. \\
\end{promptbox}

\end{document}